\pdfoutput=1

\documentclass[11pt]{article}

\usepackage[]{acl}

\usepackage{times}
\usepackage{latexsym}
\usepackage{booktabs}
\usepackage{graphicx}
\usepackage{multirow}
\usepackage{float}
\usepackage{amssymb}  
\usepackage{enumitem}
\usepackage{amsmath}
\usepackage{siunitx}
\usepackage[most]{tcolorbox}
\usepackage{listings}
\usepackage{tabularx}
\tcbuselibrary{breakable}
\newcommand{\deltaf}[1]{\hspace{2pt}\footnotesize\color{red}(#1)}

\usepackage[T1]{fontenc}

\usepackage[utf8]{inputenc}

\usepackage{microtype}

\title{From Recall to Forgetting: \\Benchmarking Long-Term Memory for Personalized Agents}

\author{
  \textbf{Md Nayem Uddin}$^{1,2}$\thanks{~~Work done during an internship at Genies.}  
  \hspace{9pt}
  \textbf{Kumar Shubham}$^{2}$ \\
  \hspace{9pt}
  \textbf{Eduardo Blanco}$^{3}$ 
  \hspace{9pt}
  \textbf{Chitta Baral}$^{1}$ 
  \hspace{9pt}
  \textbf{Gengyu Wang}$^{2}$ \\
  \hspace{9pt}
  $^{1}$Arizona State University \quad $^{2}$Genies \quad $^{3}$University of Arizona \\
  \texttt{mnuddin1@asu.edu, gwang@genies.com}
}
\begin{document}
\maketitle
\begin{abstract}

Personalized agents that interact with users over long periods must maintain persistent memory across sessions 
and update it as circumstances change.
However, existing benchmarks predominantly frame long-term memory evaluation as fact retrieval from past conversations, 
providing limited insight 
into agents’ ability to consolidate memory over time
or handle frequent knowledge updates.
We introduce Memora, 
a long-term memory benchmark spanning 
weeks to months long user conversations.
The benchmark evaluates three memory-grounded tasks: remembering, reasoning, and recommending.
To ensure data quality, we employ automated memory-grounding checks and human evaluation.
We further introduce Forgetting-Aware Memory Accuracy (FAMA), a metric that 
penalizes reliance on obsolete or invalidated memory when evaluating long-term memory. 
Evaluations of four LLMs 
and six memory agents reveal frequent reuse of invalid memories and failures to 
reconcile evolving memories.
Memory agents offer marginal improvements, exposing shortcomings in long-term memory for personalized agents.
\end{abstract}

\section{Introduction}
\label{introduction}

\begin{table}[!t]
    \centering
    \resizebox{\columnwidth}{!}{

\begin{tabular}{lrrrr}
\toprule
\multirow{3}{*}{\shortstack[l]{\textbf{Long-Term} \\ \textbf{Memory} \\ \textbf{Benchmark}}} & \multicolumn{2}{c}{\textbf{Memory}} & \multicolumn{2}{c}{\textbf{Memory}} \\
& \multicolumn{2}{c}{\textbf{Consolidation}} & \multicolumn{2}{c}{\textbf{Mutation}} \\
\cmidrule(lr){2-3} \cmidrule(lr){4-5}
& \textbf{Avg.} & \textbf{Max.} & \textbf{Avg.} & \textbf{Max.} \\
\midrule
PerLTQA~(\citeyear{du-etal-2024-perltqa}) & 1.0 & 1.0 & 0.0 & 0.0 \\
MemDaily~(\citeyear{zhang2024memsim}) & 3.3 & 8.0 & 0.0 & 0.0 \\
LOCOMO~(\citeyear{maharana-etal-2024-evaluating}) & 1.3 & 15.0 & 0.0 & 0.0 \\
LongMemEval~(\citeyear{wu2024longmemeval}) & 1.9 & 6.0 & 2.0 & 2.0 \\
PersonaMem~(\citeyear{jiang2025know}) & 1.3 & 3.0 & 1.2 & 3.0 \\
\midrule
Memora (Weekly) & 5.3 & 26.0 & 2.7 & 11.0 \\
Memora (Monthly) & 17.3 & 99.0 & 8.8 & 43.0 \\
Memora (Quarterly) & 28.4 & 309.0 & 14.8 & 94.0 \\
\bottomrule
\end{tabular}
    }
    \caption{
    Comparison of long-term memory benchmarks on memory consolidation and mutation.
\emph{Memory consolidation} measures the number of prior sessions that must be considered to answer a query, and \emph{memory mutation} measures the number of updates or deletions applied across sessions before querying. We report both average (Avg.) and maximum (Max.) values for multiple existing benchmarks. Memora introduces substantially higher consolidation and mutation requirements across weekly, monthly, and quarterly settings.
    }
    \label{tab:bench_comp}
\end{table}

\renewcommand{\thefootnote}{}
\footnotetext[1]{Our code and data are available at: \url{https://github.com/geniesinc/Memora}}
\renewcommand{\thefootnote}{\arabic{footnote}}

Large Language Models (LLMs) have rapidly advanced 
as general-purpose agents, 
demonstrating strong capabilities in 
reasoning~\cite{huang-chang-2023-towards}, 
instruction following~\cite{xu2023wizardlm, wen2024benchmarking}, 
generating high-quality content~\cite{liang2024controllable}, 
and adapting across diverse tasks~\cite{radford2019language, kojima2022large}. 
These advances have fueled growing interest in deploying LLMs as personalized assistants~\cite{yuan-etal-2025-personalized}, 
tutors~\cite{chen2024gptutorgreatpersonalizedtutor}, 
and life-long companions~\cite{zhang2025riseaicompanionshumanchatbot}. 
However, despite their apparent fluency, current LLMs remain fundamentally constrained due to the lack of persistent long-term memory~\cite{zhong2023memorybankenhancinglargelanguage, wu2025humanmemoryaimemory}. 
By default, LLMs are stateless across interactions~\cite{mei2025survey}. 
Although models maintain a key-value cache
during a single interaction to preserve short-term context, 
this internal state is discarded once the interaction ends. 
As a result, information shared by users in previous conversations, 
such as preferences, corrections, or goals is not retained unless it is explicitly reintroduced. 
This limitation prevents LLMs from behaving as persistent assistants 
that can maintain interaction over days, weeks, or months.

\begin{figure*}[t]
    \centering
    \includegraphics[width=\textwidth]{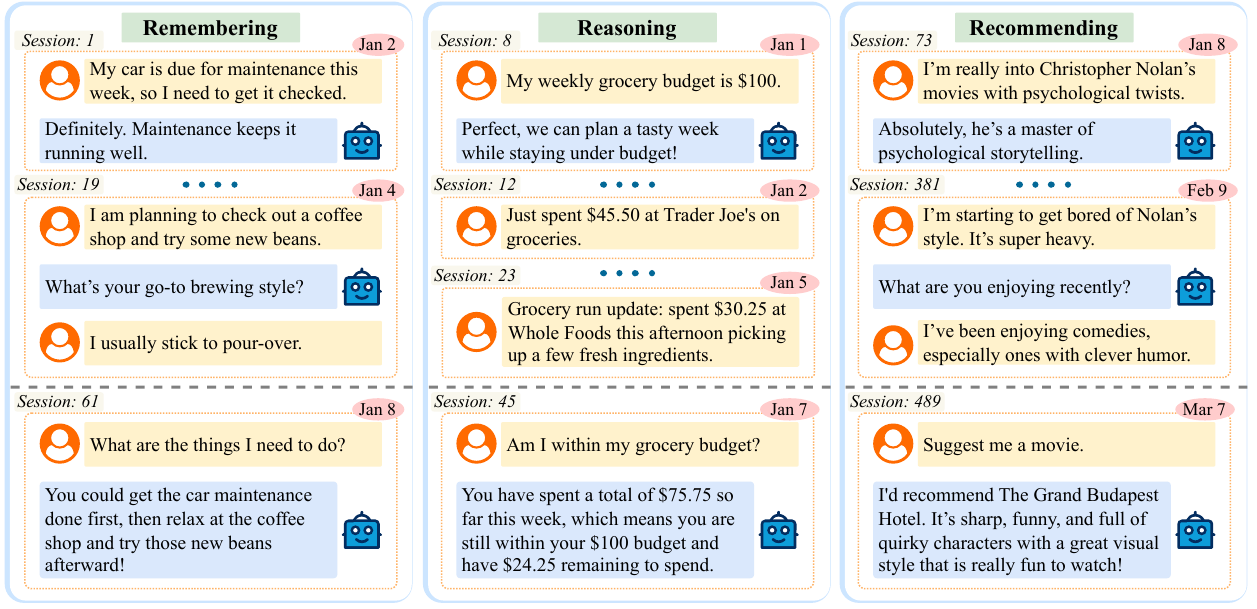}
    \caption{The three tasks of the Memora benchmark: 
    1) \textbf{Remembering:} recalling and leveraging previously discussed facts,
    such as to-dos, 
    2) \textbf{Reasoning:} integrating multiple pieces of information to derive a specific answer, for example, calculating the grocery budget status, and 
    3) \textbf{Recommending:} suggesting relevant items or actions based on the user's evolving preferences, like proposing The Grand Budapest Hotel after the user grew bored of Christopher Nolan's movies. 
    Each task depends on selectively extracting 
    and reusing relevant information from non-contiguous, temporally distant sessions, emphasizing long-term memory 
    beyond recent context.
    }
    
    \label{fig:memora_motivation} 
\end{figure*}


Human cognition provides a clear contrast. 
People naturally remember prior conversations~\cite{brown2025remembering}, 
integrate information across time~\cite{mazurek2003role}, 
revise beliefs when new evidence arises~\cite{HOGARTH19921}, 
and discard outdated knowledge~\cite{bekinschtein2018retrieval, ye2020retrieval}. 
Long-term memory is not defined solely by recalling~\cite{ericsson1995long}, 
but by the ability to accumulate experiences~\cite{meeter2004consolidation}, 
reconcile changes~\cite{wood2012review}, 
and maintain a coherent mental model of the world~\cite{jones2011mental}. 
For conversational agents to approximate this behavior, they must support not only remembering past information, 
but also consolidating memory across many interactions and mutating memory as circumstances evolve.

Despite the growing interest in long-term memory, 
existing benchmarks~\cite{maharana-etal-2024-evaluating, du-etal-2024-perltqa, jiang2025know} 
primarily operationalize it as shallow cross-session retrieval rather than sustained memory accumulation.
In LoCoMo~\cite{maharana-etal-2024-evaluating}, 94\% of the evaluation questions require grounding evidence from no more than two previous sessions. 
We observe the same pattern for 85\% of the evaluation questions in LongMemEval~\cite{wu2024longmemeval}. 
Consistent with these observations, 
Table \ref{tab:bench_comp} shows that average memory consolidation across benchmarks is 
approximately one session. 
This skewed distribution reduces most evaluations to whether a model can recall 
an isolated piece of information introduced in a prior session,
rather than synthesizing information accumulated over extended interaction histories. 
Also, this retrieval-centric framing implicitly assumes 
that stored information remains permanently valid. 
In contrast, real-world long-term interaction is non-stationary: user information is updated, corrected, or withdrawn over time. 
Therefore, long-term memory requires not only recalling past information, but also correct handling of memory mutation.
However, as shown in
Table~\ref{tab:bench_comp}, 
prior benchmarks place minimal stress on memory mutation. LongMemEval~\cite{wu2024longmemeval} includes knowledge-update operations, but limits them to at most two sessions before evaluation, 
and PersonaMem~\cite{jiang2025know} handles updates across no more than three sessions. 
As a result, models are rarely required to reconcile multiple revisions of the same information or to track 
how user states evolve over extended timelines.

To address these gaps, 
we introduce Memora, a benchmark that models long-term memory as a continuous and evolving process rather than static retrieval. 
Memora increases demands on both memory consolidation and mutation by requiring models to integrate information across weekly, monthly, and quarterly conversation sessions. 
Figure~\ref{fig:memora_motivation} shows that Memora evaluates three memory-grounded tasks: remembering, reasoning, and recommending. 
All tasks require adhering to the temporal validity of users' long-term memory.

Beyond benchmark design, 
Memora also revisits how long-term memory should be evaluated. 
Existing evaluations largely reward memory inclusion, measuring whether relevant information appears in a model’s response. 
This overlooks memory misuse, where obsolete information is retrieved and used. 
As long as the final answer appears correct, reliance on invalidated memory is not penalized.
To address this, we introduce Forgetting-Aware Memory Accuracy (FAMA), an evaluation metric that explicitly accounts for invalid memories. 
FAMA measures whether a model’s response reflects the user’s current memory state by rewarding correct use of valid memory and penalizing reliance on obsolete or deleted memory. 
This enables evaluation of memory mutation over long interaction histories.
Using Memora, we evaluate four LLMs and six long-term memory agents. 
Despite extended context windows and external memory mechanisms, our results reveal persistent failures in maintaining consistent belief states under high consolidation and mutation pressure. 
Models frequently reuse obsolete information, and long-term memory agents offer only limited improvements.
In summary, our main contributions are:
\begin{itemize}[nosep]
    \item Introducing Memora, a benchmark that substantially increases demands on both memory consolidation and memory mutation 
    across weekly, monthly and 
    quarterly durations.
    \item Proposing Forgetting-Aware Memory Accuracy (FAMA), an evaluation metric that penalizes reliance on outdated memories.
    \item Empirical evaluation of LLMs and long-term memory agents, revealing limitations in maintaining consistent memory states.
\end{itemize}
These contributions position Memora as a rigorous benchmark for studying long-term memory.
By jointly stressing memory consolidation, frequent memory mutation, and forgetting-aware evaluation, 
Memora exposes failure modes that remain hidden under retrieval-centric benchmarks.
\section{Related Works}

Long-term memory addresses 
a fundamentally different 
problem than long-context modeling~\cite{bai-etal-2024-longbench, zhang-etal-2024-bench, hsieh2024rulerwhatsrealcontext}. 
In realistic settings, placing the entire interaction history into the 
prompt is impractical~\cite{lewis2020retrieval, packer2023memgpt}
and often counterproductive~\cite{liu-etal-2024-lost, du-etal-2025-context}. 
Effective agents~\cite{park2023generativeagentsinteractivesimulacra} 
must depend on 
persistent 
and updatable long-term memory mechanisms, 
rather than simply increasing context length.

Early long-term conversational 
memory benchmarks relied on limited session 
histories~\cite{xu-etal-2022-beyond}. 
As context windows expanded, 
later benchmarks primarily emphasized scaling conversation length and explicit memory probing, including targeted recall of personal facts~\cite{zhong2024memorybank, du-etal-2024-perltqa}, 
question answering and summarization over 
long multi-session dialogues~\cite{maharana-etal-2024-evaluating}, 
narrative-driven recall in tv-series dialogues~\cite{kim2025dialsimdialoguesimulatorevaluating}, and million-tokens long user–assistant conversations~\cite{wu2024longmemeval}.

In parallel, another line of work frames 
long-term memory primarily as personalization, aiming to adapt 
agents' behavior to the individual 
users over extended interactions.
Early benchmarks such as DuLeMon~\cite{xu-etal-2022-long} evaluate persona-consistent dialogue generation. 
PersonaMem~\cite{jiang2025know} shifts toward personalized decision-making 
by testing whether models can infer evolving user states from long histories using 
multiple-choice questions.
MemDaily~\cite{zhang2024memsim} models 
daily life personal assistant interactions and probes user-specific facts and events. MemoryAgentBench~\cite{hu2025evaluatingmemoryllmagents} extends personalization-oriented memory evaluation to agentic settings, highlighting competencies such as retrieval, test-time learning, and forgetting.

Taken together, prior works have expanded 
the scale and scope of long-term memory evaluation, either by increasing conversation length or by framing memory as personalization. 
However, across both lines of work, 
long-term memory is still predominantly operationalized as 
fact-retrieval 
from past interactions, with relatively limited emphasis on memory consolidation and frequent memory mutation. As a result, it remains unclear how well existing agents integrate information across extended timelines or handle evolving and invalidated memory. 
Memora targets these challenges by jointly 
stressing consolidation and mutation in long-term memory evaluation.
\section{Memora}



Memora is constructed through a simulation-driven pipeline that jointly generates long-term conversations and evaluation tasks.
Starting from persona-level seed data, 
the pipeline simulates user interactions spanning weeks to months, 
converts these interactions into multi-turn conversations, 
and derives memory-grounded evaluation tasks.
This design focuses on 
both memory consolidation and memory mutation, 
requiring models to adhere to the temporal validity of 
information 
across the Remembering, Reasoning, and Recommending tasks.

\subsection{Seed Data Design}

We construct ten professional persona profiles (\emph{e.g., software engineers, researchers, designers, executives}) 
consisting of preference patterns, 
activity tendencies, and long-term goals. 
These personas serve as the semantic backbone of the benchmark.
Memora models three user-centric memory types: 
\emph{preference memory, activity memory, and goal memory}. 
Preference memory captures users' evolving likes and dislikes across 
domains~(\emph{e.g., movie, music, travel}). 
Activity memory represents what users' do over time, 
encompassing both 
personal activities~(\emph{e.g., expenses, fitness tracking, tasks}) 
and professional activities~(\emph{e.g., drafting documents, managing meeting notes}). 
Goal memory encodes users'
long-term objectives~(\emph{e.g., budgeting, fitness targets}).
Memory evolution is controlled 
by operational and temporal 
constraints that ensure chronological consistency across sessions. 
Further details are provided in 
Appendix \ref{app:seed_data}.

\subsection{Session Simulation}
Given the seed data, a session simulator generates  
sequences of user interactions spanning weeks to months. 
The seed data defines the space of possible memory entities,
and the simulator determines when and how those entities are 
introduced, updated, or invalidated under explicit temporal and operational constraints. The simulator also includes memory-neutral sessions 
that do not introduce, modify, or delete any stored 
memories~(\emph{e.g., casual conversations, clarifications}). 
This mixture follows interaction patterns observed in prior conversational benchmarks~\cite{wu2024longmemeval, deshpande-etal-2025-multichallenge}.

The simulator maintains a persistent memory state that is 
updated after every session. 
This enables dynamics such as 
preference drift (\emph{e.g., gradually losing interest in a favored director}), 
recurring activities (\emph{e.g., repeated activities logging}), 
and incremental progression of long-term tasks (\emph{e.g., refining a draft document across multiple sessions}).
By recording the full memory state before and after each session, 
Memora produces explicit memory traces that precisely track how information is introduced, updated, and invalidated over time. These traces define the ground truth for downstream conversation generation and memory evaluation.


\begin{table}[t]
    \centering
    \small
    \resizebox{\columnwidth}{!}{
        \normalsize
            \begin{tabular}{lrrr}
        \toprule
         & \textbf{Weekly} & \textbf{Monthly} & \textbf{Quarterly} \\
        \midrule
        Number of Personas                               & 10              & 10               & 10 \\
        Avg. Sessions Per Persona                & 155             & 615              & 1991 \\
        Avg. Turns Per Session                   & 16.1            & 15.6             & 15.7 \\
        Avg. Memory Operations       & 103.2           & 374.3            & 1171.4 \\
        \quad -- Add (\%)       & 68    & 63     & 63 \\
        \quad -- Update (\%)    & 13    & 16     & 18 \\
        \quad -- Delete (\%)    & 19    & 21     & 19  \\
        Memory-grounded Questions              & 150             & 150              & 300 \\
        Evaluation Criteria                    & 749             & 1421             & 4884 \\
        
        \midrule
        Avg. pairwise 1-gram overlap $\downarrow$ & 0.144 & 0.144 & 0.126 \\
        Avg. pairwise 2-gram overlap $\downarrow$ & 0.027 & 0.026 & 0.027 \\
        Avg. pairwise 3-gram overlap $\downarrow$ & 0.011 & 0.011 & 0.010 \\
        Avg. SBERT cosine similarity $\downarrow$ & 0.275 & 0.272 & 0.281 \\
        \bottomrule
    \end{tabular}
    }
    \caption{Memora statistics and conversation diversity across different temporal durations. The top block summarizes the benchmark scale and memory dynamics. The bottom block reports conversation diversity using 
    pairwise lexical overlap 
    and semantic similarity.}
    \label{tab:merged_data_overview}
\end{table}

\begin{figure*}[hbt!]
    \centering
    \includegraphics[width=\textwidth]{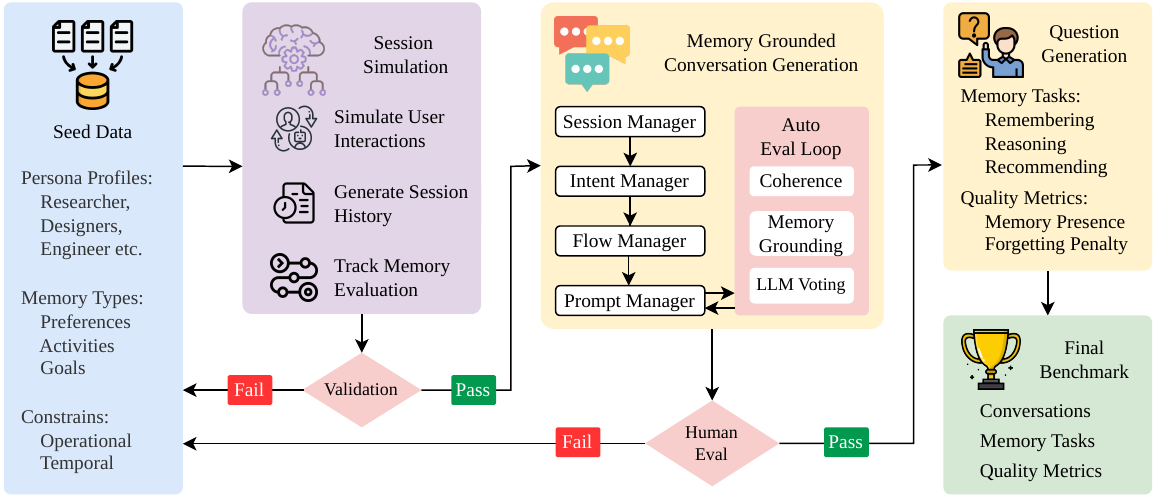}
    \caption{
    Overview of the Memora construction pipeline. 
    The process begins with structured 
    seed data (persona profiles, memory types, constraints) that drives the session simulation module to produce long-term interaction histories. 
    Conversations are generated by multiple LLM agents. 
    An auto-eval loop checks coherence and memory grounding. 
    Rigorous validation checkpoints, 
    including both internal mechanisms (LLM Voting) and external human evaluation, filter the generated data 
    for quality and correctness before 
    forming the final benchmark.}
    \label{fig:memora_flow} 
\end{figure*}

\subsection{Conversation Generation}

Building on the simulated session history, 
Memora converts each session specification 
into a multi-turn dialogue 
using a controlled, multi-agent conversation generation framework. 
The framework supports two types of conversational turns: 
1) memory-neutral turns, which involve general turns (\emph{e.g., casual questions, acknowledgments}), 
and 2) memory-grounded turns, in which the user expresses information corresponding to the simulated memory operation.
An intent selection module determines a sequence of user and assistant intents (\emph{e.g., greeting, memory disclosure, follow-up}).
Conversations are organized into a multi-phase interaction consisting of an opening phase, an exploration phase, a memory phase where the target operation is expressed, and a closing phase. Dialogue turns are generated using a multi-agent prompting setup with separate user and assistant roles conditioned on persona traits, selected intents, memory entities, and prior conversation context.

LLM-based generation does not always strictly adhere to instructions given in the prompt and may introduce plausible but untracked memory details beyond the simulated session specification. 
To address this, all generated conversations are checked through 
an automated memory-grounding evaluation loop.
The grounding checks verify that the intended memory operation and the entity 
is correctly expressed in the conversation, 
and no additional information is introduced. 
Each conversation is independently evaluated by three LLMs 
and accepted only if all agree. 
Otherwise feedback is shared and the conversation is regenerated. 
This iterative process promotes 
close alignment between generated conversations 
and the underlying memory trace. 
In addition to automated checks, we randomly sample 5\% of generated conversations per persona for human verification. If annotators identify any inconsistency between the conversation and the memory trace, the entire batch is rejected.
Further details are provided in Appendix \ref{app:conversation_gen}.

\subsection{Conversation Diversity in Memora}
Table \ref{tab:merged_data_overview} summarizes conversation 
scale and linguistic diversity across weekly, monthly, and quarterly timelines. Generated conversations exhibit low average pairwise n-gram overlap, 
indicating minimal template reuse 
and broad lexical coverage. 
Low average SBERT cosine similarity  
shows that diversity extends beyond surface form to semantic content. 
Together, these results demonstrate that Memora generates 
linguistically diverse conversations without collapsing into formulaic patterns.

\subsection{Questions and Evaluation Criteria}
Memora constructs evaluation questions directly from the simulated memory traces
(structured record of all memory states
and their updates across sessions).
Questions are organized into three tasks: \emph{Remembering}, \emph{Reasoning}, and \emph{Recommending}. 
Remembering questions test direct recall of 
stored information (\emph{e.g., generating documents}), 
Reasoning questions require synthesizing 
information~(\emph{e.g., evaluating goal progress}), and 
Recommending questions assess whether personalized suggestions 
reflect the user’s current preferences rather than outdated ones 
(\emph{e.g., recommending a movie after preference changes}). 
Each question is paired with explicit evaluation criteria derived from the memory trace, consisting of (i) \emph{memory presence} criteria that specify which valid information must be included in the response, and (ii) \emph{forgetting absence} criteria that specify which outdated or invalidated information must be excluded.


\subsection{Final Benchmark}

The Memora benchmark consists of validated multi-session conversations, memory-grounded evaluation questions, and evaluation criteria anchored in explicit memory traces for each persona. Correct responses require integrating information across multiple sessions while avoiding invalidated memories,
enabling fine-grained analysis of long-term memory beyond surface-level accuracy. 
Example samples are provided in Appendix~\ref{app:conversation_examples}.
\section{Experiments}
We evaluate long-term memory behavior on Memora under two settings: 
1) large language models operating directly over conversational histories, and 
2) long-term memory agents that explicitly store and retrieve user information across sessions. These settings isolate different mechanisms for maintaining memory and 
allow us to assess whether evaluated models and agents produce consistent responses with the user’s memory state.

\subsection{Evaluation Settings}
\textbf{Model-Based Evaluation}: 
We evaluate LLMs by providing multi-session conversation histories~(as permitted by the context window) and asking them to answer memory-grounded questions. 
This setting isolates models’ ability to consolidate long interaction histories without external memory modules. We evaluate four LLMs: GPT-5.2, Claude Sonnet 4.5, Gemini 3 Pro Preview, and Qwen3-32B. They are tested under both standard and reasoning-enabled inference to assess whether reasoning tokens improves memory under frequent updates.

\noindent\textbf{Agent-Based Evaluation}: We evaluate long-term memory agents that 
incrementally ingest prior conversations, retrieve relevant memories at query time, and generate responses conditioned on retrieved memories. We include representative memory agents spanning local vector stores, cloud-based memory APIs, profile-driven memories, and stateful agents, evaluated under identical conversations.
The long-term memory agents are A-Mem, LangMem, Mem-0, MemoBase, MemoryOS, Nemori. 
All agents use the same LLM (GPT-4o-mini) 
backend for answer generation.

\subsection{Forgetting-Aware Memory Accuracy}
Memora evaluates responses using atomic, 
memory-aligned criteria derived from 
the user's memory state. 
Each evaluation question is paired with two groups of binary criteria:
\textit{memory presence} criteria, which check whether 
valid information is correctly included in the response, 
and \textit{forgetting absence} criteria, which check whether 
invalidated or deleted information is properly excluded. 
This distinction separates correct reliance on the memory 
from the erroneous reuse of obsolete memory, which 
standard accuracy metrics do not capture.


Each criterion is evaluated independently using LLM-based judges. Given a model response and a single criterion, three judges—GPT-4.1, Claude Haiku 4.5, and Gemini 2.5 Flash—each provide a binary decision (“yes” or “no”). The final outcome is determined by majority vote.
This evaluation setup follows prior work on LLM-as-judge methods for open-ended and long-context tasks~\cite{bai-etal-2024-longbench, maharana-etal-2024-evaluating, es-etal-2024-ragas}. To validate reliability, we conduct a human evaluation study, which shows an average agreement of 88.3\% between LLM judgments and human annotations. Inter-annotator agreement, measured using Cohen’s $\kappa$, ranges from 0.86 to 0.90.
Additional details are provided in Appendix~\ref{app:add_eval}.

We introduce Forgetting-Aware Memory Accuracy (FAMA) to aggregate criterion-level judgments into a single score.
FAMA rewards correct use of valid memory while explicitly 
penalizing reliance on obsolete memory.
\begingroup
\setlength{\abovedisplayskip}{2pt}
\setlength{\belowdisplayskip}{2pt}
\[
\text{FAMA} = \max\!\Big(0,\; \text{MPA} - \lambda \cdot (1 - \text{FAA})\Big)
\]
\endgroup
where MPA (memory presence accuracy) is the fraction of memory presence 
criteria satisfied, and FAA (forgetting absence accuracy) is the fraction 
of forgetting absence criteria satisfied. 
The weighting term $\lambda$ is defined per question as follows:
\begingroup
\setlength{\abovedisplayskip}{2pt}
\setlength{\belowdisplayskip}{2pt}
\[
\lambda = \frac{N_{\text{forget}}}{N_{\text{presence}} + N_{\text{forget}}},
\]
\endgroup
where $N_{\text{presence}}$ and $N_{\text{forget}}$ are the number of 
memory presence and forgetting absence criteria for that question.
The $\max$ operator ensures 
FAMA remains non-negative. 
Per-question FAMA is thus bounded in $[0, 1]$.
For each of the three tasks 
(Remembering, Reasoning, Recommending), 
we sum per-question FAMA scores across all questions 
within that task and normalize to $[0, 100]$. 
Table~\ref{tab:main_results} reports these task-level scores 
individually for each temporal duration, 
enabling direct comparison across tasks and timelines.
\section{Results}

\begin{table*}[ht]
    \centering
    
    \resizebox{0.93\textwidth}{!}{%

\begin{tabular}{l rrr | rrr | rrr}
    \toprule
    \multirow{2}{*}{\textbf{Models / Agents}} & 
    \multicolumn{3}{c}{\textbf{Remembering}} & 
    \multicolumn{3}{c}{\textbf{Recommending}} & 
    \multicolumn{3}{c}{\textbf{Reasoning}} \\
    
    \cmidrule(lr){2-4} \cmidrule(lr){5-7} \cmidrule(lr){8-10}
    
     & \textbf{Weekly} & \textbf{Monthly} & \textbf{Quarterly} & 
       \textbf{Weekly} & \textbf{Monthly} & \textbf{Quarterly} & 
       \textbf{Weekly} & \textbf{Monthly} & \textbf{Quarterly} \\ 
    \midrule
    
    \multicolumn{10}{l}{\emph{Language Models (w/o Reasoning Tokens)}} \\ 
    \hspace{3mm} Qwen3-32B                      & 26.12 & 21.14 & 19.24    & 50.16 & 50.30 & 48.88   & 6.00 & 2.00 & 6.00 \\
    \hspace{3mm} Claude Sonnet 4.5              & 27.50 & 19.42 & 21.25    & 43.62 & 39.00 & 44.02   & 6.66 & 3.00 & 5.50 \\
    \hspace{3mm} Gemini 3 Pro Preview           & 20.36 & 21.44 & 17.28    & 45.12 & 45.94 & 52.56   & 6.66 & 4.00 & 4.00 \\
    \hspace{3mm} GPT-5.2                        & 25.32 & 19.92 & 23.39    & 54.80 & 51.12 & 53.36   & 4.66 & 0.00 & 1.00 \\
    \midrule
    
    \multicolumn{10}{l}{\emph{Language Models (w/ Reasoning Tokens)}} \\
    \hspace{3mm} Qwen3-32B                      & 23.86 & 25.62 & 17.14    & 50.04 & 53.06 & 47.71   & 6.66 & 9.00 & 3.00 \\
    \hspace{3mm} Claude Sonnet 4.5              & 26.56 & 21.40 & 19.13    & 52.40 & 60.90 & 51.78   & 4.00 & 0.00 & 2.50 \\
    \hspace{3mm} Gemini 3 Pro Preview           & 21.02 & 23.26 & 18.12    & 43.36 & 44.92 & 50.83   & 6.00 & 10.00 & 8.50 \\
    \hspace{3mm} GPT-5.2                        & 25.70 & 19.22 & 22.16    & 53.40 & 51.60 & 53.36   & 4.66 & 0.00 & 2.00 \\
    \midrule
    
    \multicolumn{10}{l}{\emph{Long-Term Memory Agents}} \\
    \hspace{3mm} A-Mem                          & 71.82 & 41.90 & 40.78     & 35.04 & 37.52 & 34.95   & 2.00  & 2.00 & 5.00 \\
    \hspace{3mm} LangMem                        & 71.16 & 42.00 & 39.14     & 48.88 & 44.08 & 33.85   & 30.00 & 14.00 & 11.00 \\
    \hspace{3mm} Mem-0                          & 40.42 & 21.08 & 19.90     & 52.58 & 36.20 & 38.47   & 16.00 & 0.00 & 2.00 \\
    \hspace{3mm} MemoBase                       & 43.60 & 20.08 & 15.18     & 68.94 & 58.46 & 45.62   & 18.00 & 7.00 & 1.00 \\
    \hspace{3mm} MemoryOS                       & 51.84 & 29.78 & 25.05     & 62.64 & 48.54 & 44.02   & 20.66 & 6.00 & 5.50 \\
    \hspace{3mm} Nemori                         & 65.06 & 44.08 & 33.83     & 52.84 & 45.90 & 41.66   & 18.66 & 0.00 & 6.50 \\
    
    \bottomrule
\end{tabular}
    }
    \caption{
    Task-level FAMA scores aggregated over evaluation questions.
    For each task (remembering, recommending, reasoning) and temporal duration (weekly, monthly, quarterly), scores are computed by summing per-question FAMA scores and normalizing to [0, 100]. This represents performance within each task–duration setting. 
    }
    \label{tab:main_results}
\end{table*}

We analyze Forgetting-Aware Memory Accuracy (FAMA) scores to understand how LLMs and long-term memory agents behave under increasing temporal span, memory mutation requirements, and task complexity. Overall, three patterns emerge. First, performance generally declines from weekly to quarterly settings, showing that longer and more mutation-heavy interaction histories make memory use less reliable. Second, performance is strongly task-dependent: long-term memory agents are strongest on remembering, language models remain competitive on recommending, and reasoning is difficult for all LLMs and agents. Third, forgetting-aware evaluation reveals substantial reliance on outdated or invalidated memory that standard memory accuracy does not capture.

\noindent\textbf{Performance Across Temporal Durations:}
A clear pattern in Table~\ref{tab:main_results} is that performance generally degrades from the weekly to the quarterly setting. 
As the temporal horizon expands, agents must operate over substantially longer conversation histories with more accumulated memories, more updates, and more invalidations. 
This makes maintaining a temporally correct memory state 
increasingly difficult. 
The week-to-quarter degradation is most consistent in \textit{remembering}: all LLMs and agents perform worse in the quarterly setting than in the weekly setting. This drop is especially pronounced for memory agents, including MemoBase (43.6 to 15.18), MemoryOS (51.84 to 25.05), and Mem-0 (40.42 to 19.90). This shows that even agents with explicit memory stores become increasingly brittle as memory grows longer and more mutation-heavy.
The same weekly-to-quarterly decline also appears in most cases for \textit{reasoning}, where 11 of 14 LLMs and agents perform worse at quarterly scale. Here, the degradation is particularly severe because reasoning already starts from a low baseline: for example, MemoBase (18.00 to 1.00) and MemoryOS (20.66 to 5.50).

\begin{table}[!ht]
\centering
\small
\resizebox{\columnwidth}{!}{%
\begin{tabular}{lccc}
\toprule
\textbf{Models / Agents} & \textbf{Remembering} & \textbf{Recommending} & \textbf{Reasoning} \\ 
\midrule
\multicolumn{4}{l}{\textit{Language Models (w/o Reasoning Tokens)}} \\
Qwen3-32B & 66.50 & 149.34 & 14.00 \\
Claude Sonnet 4.5 & 68.17 & 126.64 & 15.16 \\
Gemini 3 Pro P. & 59.08 & 143.62 & 14.66 \\
GPT-5.2 & 68.63 & 159.28 & 5.66 \\
\midrule
\textbf{Average} & \textbf{65.60} & \textbf{144.72} & \textbf{12.37} \\
\midrule
\multicolumn{4}{l}{\textit{Language Models (w/ Reasoning Tokens)}} \\
Qwen3-32B & 66.62 & 150.81 & 18.00 \\
Claude Sonnet 4.5 & 67.09 & 165.08 & 6.50 \\
Gemini 3 Pro P. & 62.40 & 139.11 & 24.50 \\
GPT-5.2 & 67.08 & 158.36 & 6.66 \\
\midrule
\textbf{Average} & \textbf{65.80} & \textbf{153.34} & \textbf{13.92} \\
\midrule
\multicolumn{4}{l}{\textit{Long-Term Memory Agents}} \\
A-Mem & 154.50 & 107.51 & 9.00 \\
LangMem & 152.30 & 126.81 & 55.00 \\
Mem-0 & 81.40 & 127.25 & 18.00 \\
MemoBase & 78.86 & 173.02 & 26.00 \\
MemoryOS & 106.67 & 155.20 & 32.16 \\
Nemori & 142.97 & 140.40 & 25.16 \\
\midrule
\textbf{Average} & \textbf{119.45} & \textbf{138.37} & \textbf{27.55} \\
\bottomrule
\end{tabular}%
}
\caption{Aggregated FAMA obtained by summing the task-level scores from Table \ref{tab:main_results} across weekly, monthly, and quarterly durations. This aggregation highlights clear task-dependent differences: long-term memory agents outperform LLMs on remembering, language models remain competitive with top agents on recommending, and performance on reasoning remains low for both LLMs and long-term memory agents.}
\label{tab:task_aggregation}
\end{table}

\noindent \textit{Recommending} shows the weakest week-to-quarterly degradation. Most LLMs and agents still decline (11 of 14), 
especially memory agents such as 
MemoBase (68.94 to 45.62), 
MemoryOS (62.64 to 44.02), and 
Nemori (52.84 to 41.66). 
However, recommendation also contains most of the exceptions where performance remains stable or even improves, 
such as Gemini 3 Pro Preview and Claude Sonnet 4.5. 
We attribute this to the nature of the task. Unlike remembering and reasoning, which are evaluated against more determinate fact-based criteria derived from the current memory state, recommendation allows a wider range of acceptable responses. As a result, models can still receive credit by generating plausible suggestions that are broadly consistent with the user’s current preferences, even when explicit memory retrieval is incomplete, because they can infer likely preferences from persona-aligned conversational cues that remain available in the context window.

\noindent\textbf{Performance Across Tasks:}
Table~\ref{tab:task_aggregation} shows that long-term memory performance differs by task. Memory agents are clearly the strongest on \textit{remembering}, achieving 119.45 average aggregated FAMA versus 65.60--65.80 for language models. This large gap highlights the value of explicit memory mechanisms for factual recall.
The reasoning tokens provide little benefit for LLMs on this task.
By contrast, \textit{recommending} is the only task where language models match or exceed memory agents, with average scores of 144.72 without reasoning and 153.34 with reasoning, compared to 138.37 for agents. \textit{Reasoning} is the weakest task overall. Even the best average performance remains low, at 27.55 for memory agents, compared to 12.37 and 13.92 for language models. These results show that long-term memory is not a single capability: agents that retrieve well do not necessarily reason well over temporally distributed memory.

\noindent\textbf{The effect of Forgetting-Aware Evaluation:} 
Standard memory evaluation based 
on memory presence accuracy overestimates 
long-term memory performance.
These metrics evaluate the final response by checking whether 
required information appears, 
but they do not penalize the use of obsolete or invalidated memory. 
As a result, models can achieve high scores even when their responses 
conflate between past and current memory states.
Table~\ref{tab:memory_fama_diff} compares aggregated memory-presence accuracy scores with the forgetting-aware reductions introduced by 
the proposed FAMA metric.
Across all language models and long-term memory agents, applying forgetting-aware evaluation 
results in large score reductions.

The score reductions follow different trends for language models and memory agents. For language models, the reduction decreases as temporal range grows (from 32.6 weekly to 17.8 quarterly), not because memory improves, but because longer histories exceed the context window and relevant information is omitted altogether. Memory agents exhibit the opposite trend. Their score reductions increase with longer timelines (from 18.2 weekly to 29.5 quarterly), showing that as memory scales, agents increasingly rely on information that should have been revised or discarded. Retaining access to older memories without effective forgetting amplifies inconsistency.

Consequently, long-term memory agents with the same aggregated memory presence accuracy receive different forgetting-based reductions, leading to changes in their final performance rankings. For example, in the monthly setting, MemoryOS (112.8) and A-Mem (112.0) both outperform Nemori (105.4) under memory presence accuracy, but Nemori receives a much smaller forgetting-based reduction (15.4 vs. 28.4 and 29.5), yielding a higher final aggregated FAMA score and moving it ahead of both systems in the final ranking.

\begin{table}[t]
\centering
\small
\resizebox{\columnwidth}{!}{%
    \begin{tabular}{l *{3}{S[table-format=3.2, table-column-width=0.75cm] @{\extracolsep{0pt}} l}}
    \toprule
    \multirow{2}{*}{\textbf{Models / Agents}} & \multicolumn{6}{c}{\textbf{Aggregated Memory Presence Accuracy}} \\
    \cmidrule(lr){2-7}
     & \multicolumn{2}{c}{\textbf{Weekly}} & \multicolumn{2}{c}{\textbf{Monthly}} & \multicolumn{2}{c}{\textbf{Quarterly}} \\
    \midrule
    
    \multicolumn{7}{l}{\textit{Language Models (w/o Reasoning Tokens)}} \\ 
    Qwen3-32B          & 103.6 & \deltaf{$-21.3$} & 83.0  & \deltaf{$-9.7$}  & 79.5 & \deltaf{$-5.3$} \\
    Claude Sonnet 4.5  & 114.0 & \deltaf{$-36.2$} & 100.2 & \deltaf{$-34.1$} & 94.0  & \deltaf{$-23.2$} \\
    Gemini 3 Pro P.      & 114.2 & \deltaf{$-42.0$} & 110.6 & \deltaf{$-39.4$} & 101.8 & \deltaf{$-27.9$} \\
    GPT-5.2            & 115.4 & \deltaf{$-30.7$} & 96.6  & \deltaf{$-21.8$} & 92.5  & \deltaf{$-14.7$} \\
    \midrule
    
    \multicolumn{7}{l}{\textit{Language Models (w/ Reasoning Tokens)}} \\ 
    Qwen3-32B          & 98.7  & \deltaf{$-18.8$} & 97.8  & \deltaf{$-10.1$} & 79.5  & \deltaf{$-11.6$} \\
    Claude Sonnet 4.5  & 114.0 & \deltaf{$-31.0$} & 104.2 & \deltaf{$-21.9$} & 83.2  & \deltaf{$-9.8$} \\
    Gemini 3 Pro P.      & 113.5 & \deltaf{$-43.2$} & 111.4 & \deltaf{$-33.2$} & 95.8  & \deltaf{$-18.6$} \\
    GPT-5.2            & 111.5 & \deltaf{$-27.8$} & 97.4  & \deltaf{$-26.5$} & 91.8  & \deltaf{$-14.2$} \\
    \midrule
    
    \multicolumn{7}{l}{\textit{Long-Term Memory Agents}} \\
    A-Mem              & 118.0 & \deltaf{$-9.1$}  & 112.0 & \deltaf{$-29.5$} & 118.6 & \deltaf{$-37.9$} \\
    LangMem            & 173.0 & \deltaf{$-23.0$} & 132.2 & \deltaf{$-31.1$} & 127.4 & \deltaf{$-43.4$} \\
    Mem-0              & 119.4 & \deltaf{$-10.4$} & 78.6  & \deltaf{$-21.3$} & 72.7  & \deltaf{$-12.3$} \\
    MemoBase           & 154.4 & \deltaf{$-23.8$} & 107.2 & \deltaf{$-21.6$} & 93.7  & \deltaf{$-31.9$} \\
    MemoryOS           & 155.2 & \deltaf{$-20.6$} & 112.8 & \deltaf{$-28.4$} & 99.6  & \deltaf{$-25.0$} \\
    Nemori             & 159.4 & \deltaf{$-22.8$} & 105.4 & \deltaf{$-15.4$} & 106.8 & \deltaf{$-26.3$} \\
    \bottomrule
\end{tabular}
}
\caption{Aggregated memory-presence accuracy across all three tasks for each temporal duration. Parenthesized values show the score reduction after applying the forgetting-aware penalty in FAMA; larger reductions indicate heavier reliance on obsolete memory.}
\label{tab:memory_fama_diff}
\end{table}



\begin{figure}[ht]
    \centering
    \includegraphics[width=\columnwidth]{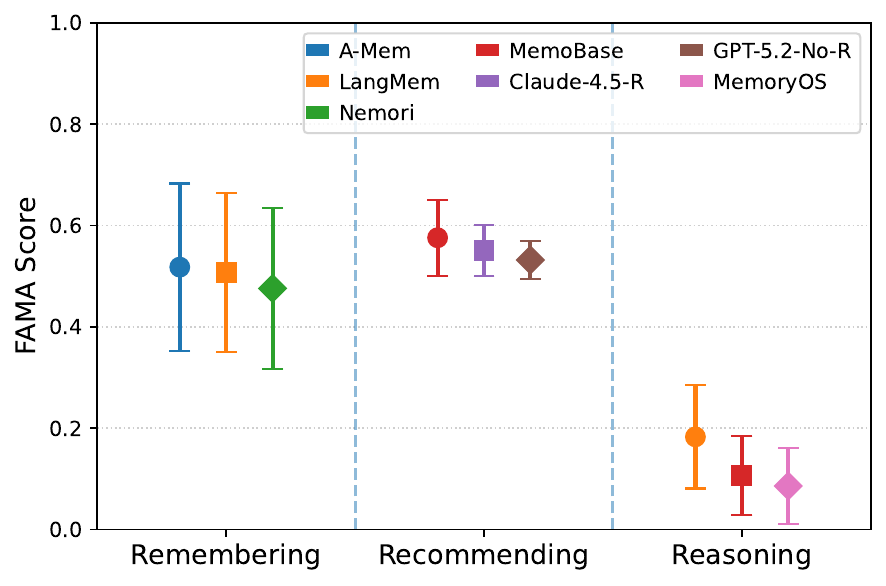}
    \caption{
    FAMA scores for remembering, recommending, and reasoning tasks. For each task, we report the top three approaches. Points denote mean per-question FAMA scores, and error bars indicate variability across temporal durations.
    }
    \label{fig:task_analysis}
\end{figure}
\begin{figure}[ht]
    \centering
    \includegraphics[width=\columnwidth]{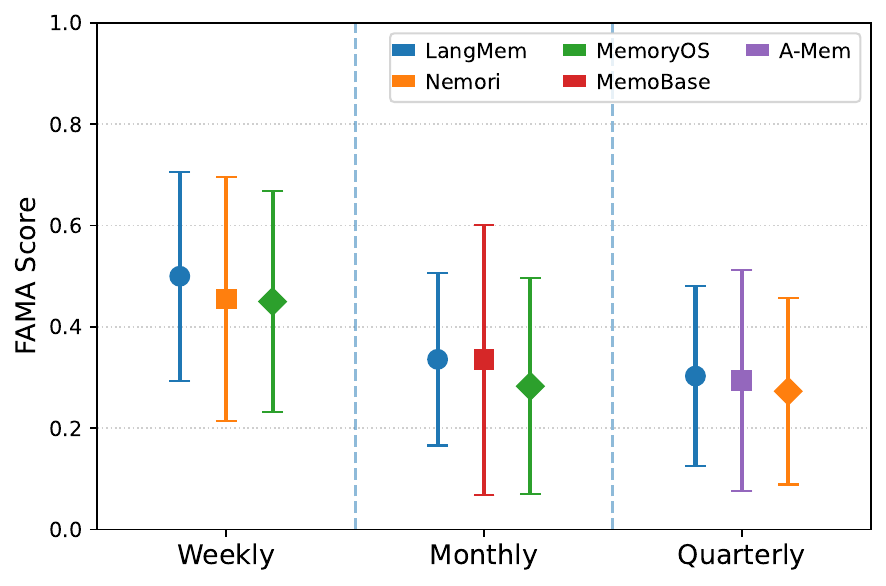}
    \caption{FAMA scores for weekly, monthly, and quarterly durations. For each temporal duration, we report the top three approaches. Points denote mean per-question FAMA scores, and error bars indicate variability across tasks.}
    \label{fig:temporal_analysis}
\end{figure}

\noindent\textbf{Forgetting-aware performance variability:} 
Figures \ref{fig:task_analysis} and \ref{fig:temporal_analysis} provide complementary views of per-question FAMA across tasks and temporal durations. Both figures reveal substantial variability, indicating that long-term memory behavior is unstable across both time and task settings. 
In Figure \ref{fig:task_analysis}, the large error bars across durations show that systems are sensitive to temporal scaling, with longer interaction histories introducing inconsistent performance. 
In Figure \ref{fig:temporal_analysis}, variability across tasks highlights that long-term memory is not a unified capability: systems that perform well on recommendation often fail on reasoning. This asymmetry reflects differing task requirements. Recommendation tolerates partial or approximate memory, allowing models to produce plausible outputs even with incomplete retrieval. In contrast, reasoning requires consistent integration of multiple memory elements, making it highly sensitive to missing or outdated information. This explains the consistently low reasoning scores across all systems.

Finally, the divergence between mean performance and variability suggests that current systems are not only inaccurate but also brittle, with performance highly sensitive to temporal and task conditions. These findings indicate that long-term memory challenges stem less from memory capacity limitations (e.g., context length or storage size) and more from failures to maintain coherent, up-to-date memory under frequent mutation.

\section{Error Analysis}

We conduct a manual error analysis over 75 incorrect predictions, randomly sampled across all temporal durations, with 25 samples per task. For each task, we analyze errors from the best-performing long-term memory agent.
Recommendation errors are primarily driven by the failures to forget outdated memory and partial memory retrieval.
We found 16 of 25 errors (64\%) 
were caused by outdated memory not being forgotten, and 7 of 25 (28\%) 
errors involved partial retrieval of preferences.
Agents often retrieve historical preferences while failing to apply recent updates. 
For example, a user initially preferred non-fiction books but later shifted toward contemporary fiction. When asked for a recommendation, the agent suggested a historical biography.
Remembering errors are dominated by partial memory retrieval: 18 of 25 errors (72\%), often resulting in incomplete structured outputs. Agents retrieve some but not all required memory items.
For example, a project summary request initially included objectives and deadlines, with collaborators and a risk assessment added later. The generated summary omitted the later-added items.
Reasoning errors consistently involve partial retrieval that prevents consolidation. 
All reasoning errors (100\%) 
involved incomplete retrieval of relevant memory elements, preventing correct consolidation.
For example, after logging multiple expenses under a monthly budget, agents responded with vague judgments (e.g., within budget) rather than computing the remaining amount.

In summary, error patterns are task-specific: recommending failures due to outdated or partial preferences, remembering failures due to incomplete retrieval, and reasoning failures because missing memory elements prevent consolidation. These highlight the need for task-aware memory mechanisms that jointly support retrieval, forgetting, and consolidation for long-term memory.
\section{Conclusion}
Memora serves as a controlled stress test that isolates key long-term memory challenges and enables more diagnostic evaluation. By grounding interactions in explicit memory traces, it assesses whether models maintain temporally consistent memory states rather than relying on isolated recall. We further introduce Forgetting-Aware Memory Accuracy (FAMA), which penalizes reliance on invalidated memory and exposes substantial performance gaps across both LLMs and long-term memory agents that standard metrics fail to capture. Together, these findings suggest that advancing long-term conversational memory will require mechanisms that explicitly integrate forgetting, consolidation, and mutation as first-class design principles.

\section*{Limitations}
Memora aims to provide a controlled and challenging benchmark for long-term conversational memory, which necessarily involves several design trade-offs. \textbf{First}, Memora relies on simulated long-horizon conversations with explicit memory creation, mutation, and deletion. 
While simulation cannot fully capture the ambiguity and unpredictability of real user interactions, collecting and manually annotating real-world memory logs over weeks or months is very costly. Such data would require user consent, 
careful privacy handling, and 
manual annotation of evolving user states, 
making it difficult to scale or standardize. Importantly, simulation does not simplify the task for evaluated systems: models already struggle under these controlled conditions. 
Since real deployments introduce additional complexities such as implicit updates and contradictory signals, 
Memora should be viewed as a lower bound. 
Systems that fail in simulation are unlikely to generalize to more complex real-world settings, making the benchmark a meaningful stress test despite its synthetic nature.
\textbf{Second}, Memora centers on a constrained set of personas and memory categories (preferences, activities, and goals) that are directly relevant to personalized assistants. This scope excludes other forms of long-term memory, such as social relationships and multi-user coordination. We leave the inclusion of richer social and relational memory structures to future work.
\textbf{Third}, evaluation relies primarily on LLM-based judges with majority voting to ensure scalability. Although automated judging may introduce shared biases, using multiple judge models and criterion-level decisions reduces variance and dependence on any single evaluator. 
\textbf{Finally}, we do not report runtime, latency, or efficiency metrics. Participating systems rely on heterogeneous hardware and storage infrastructures, making fair efficiency comparisons difficult. We therefore focus on correctness and robustness of memory usage rather than potentially misleading performance measurements.


\section*{Ethical Considerations}
The authors state that this work is in accordance
with the ACL Code of Ethics and does not raise
ethical issues. AI assistants, 
specifically Grammarly and ChatGPT, were utilized to correct grammatical errors and restructure sentences.

\bibliography{anthology,custom}

@inproceedings{zhong2024memorybank,
  title={Memorybank: Enhancing large language models with long-term memory},
  author={Zhong, Wanjun and Guo, Lianghong and Gao, Qiqi and Ye, He and Wang, Yanlin},
  booktitle={Proceedings of the AAAI Conference on Artificial Intelligence},
  volume={38},
  number={17},
  pages={19724--19731},
  year={2024}
}

@inproceedings{xu-etal-2022-beyond,
    title = "Beyond Goldfish Memory: Long-Term Open-Domain Conversation",
    author = "Xu, Jing  and
      Szlam, Arthur  and
      Weston, Jason",
    editor = "Muresan, Smaranda  and
      Nakov, Preslav  and
      Villavicencio, Aline",
    booktitle = "Proceedings of the 60th Annual Meeting of the Association for Computational Linguistics (Volume 1: Long Papers)",
    month = may,
    year = "2022",
    address = "Dublin, Ireland",
    publisher = "Association for Computational Linguistics",
    url = "https://aclanthology.org/2022.acl-long.356/",
    doi = "10.18653/v1/2022.acl-long.356",
    pages = "5180--5197"
}

@inproceedings{du-etal-2024-perltqa,
    title = "{P}er{LTQA}: A Personal Long-Term Memory Dataset for Memory Classification, Retrieval, and Fusion in Question Answering",
    author = "Du, Yiming  and
      Wang, Hongru  and
      Zhao, Zhengyi  and
      Liang, Bin  and
      Wang, Baojun  and
      Zhong, Wanjun  and
      Wang, Zezhong  and
      Wong, Kam-Fai",
    editor = "Wong, Kam-Fai  and
      Zhang, Min  and
      Xu, Ruifeng  and
      Li, Jing  and
      Wei, Zhongyu  and
      Gui, Lin  and
      Liang, Bin  and
      Zhao, Runcong",
    booktitle = "Proceedings of the 10th SIGHAN Workshop on Chinese Language Processing (SIGHAN-10)",
    month = aug,
    year = "2024",
    address = "Bangkok, Thailand",
    publisher = "Association for Computational Linguistics",
    url = "https://aclanthology.org/2024.sighan-1.18/",
    pages = "152--164"
}

@article{zhang2024memsim,
  title={Memsim: A bayesian simulator for evaluating memory of llm-based personal assistants},
  author={Zhang, Zeyu and Dai, Quanyu and Chen, Luyu and Jiang, Zeren and Li, Rui and Zhu, Jieming and Chen, Xu and Xie, Yi and Dong, Zhenhua and Wen, Ji-Rong},
  journal={arXiv preprint arXiv:2409.20163},
  year={2024}
}

@inproceedings{maharana-etal-2024-evaluating,
    title = "Evaluating Very Long-Term Conversational Memory of {LLM} Agents",
    author = "Maharana, Adyasha  and
      Lee, Dong-Ho  and
      Tulyakov, Sergey  and
      Bansal, Mohit  and
      Barbieri, Francesco  and
      Fang, Yuwei",
    editor = "Ku, Lun-Wei  and
      Martins, Andre  and
      Srikumar, Vivek",
    booktitle = "Proceedings of the 62nd Annual Meeting of the Association for Computational Linguistics (Volume 1: Long Papers)",
    month = aug,
    year = "2024",
    address = "Bangkok, Thailand",
    publisher = "Association for Computational Linguistics",
    url = "https://aclanthology.org/2024.acl-long.747/",
    doi = "10.18653/v1/2024.acl-long.747",
    pages = "13851--13870"
}

@article{wu2024longmemeval,
  title={Longmemeval: Benchmarking chat assistants on long-term interactive memory},
  author={Wu, Di and Wang, Hongwei and Yu, Wenhao and Zhang, Yuwei and Chang, Kai-Wei and Yu, Dong},
  journal={arXiv preprint arXiv:2410.10813},
  year={2024}
}

@article{jiang2025know,
  title={Know me, respond to me: Benchmarking llms for dynamic user profiling and personalized responses at scale},
  author={Jiang, Bowen and Hao, Zhuoqun and Cho, Young-Min and Li, Bryan and Yuan, Yuan and Chen, Sihao and Ungar, Lyle and Taylor, Camillo J and Roth, Dan},
  journal={arXiv preprint arXiv:2504.14225},
  year={2025}
}

@article{xu2023wizardlm,
  title={Wizardlm: Empowering large language models to follow complex instructions},
  author={Xu, Can and Sun, Qingfeng and Zheng, Kai and Geng, Xiubo and Zhao, Pu and Feng, Jiazhan and Tao, Chongyang and Jiang, Daxin},
  journal={arXiv preprint arXiv:2304.12244},
  year={2023}
}

@article{wen2024benchmarking,
  title={Benchmarking complex instruction-following with multiple constraints composition},
  author={Wen, Bosi and Ke, Pei and Gu, Xiaotao and Wu, Lindong and Huang, Hao and Zhou, Jinfeng and Li, Wenchuang and Hu, Binxin and Gao, Wendy and Xu, Jiaxing and others},
  journal={Advances in Neural Information Processing Systems},
  volume={37},
  pages={137610--137645},
  year={2024}
}

@inproceedings{huang-chang-2023-towards,
    title = "Towards Reasoning in Large Language Models: A Survey",
    author = "Huang, Jie  and
      Chang, Kevin Chen-Chuan",
    editor = "Rogers, Anna  and
      Boyd-Graber, Jordan  and
      Okazaki, Naoaki",
    booktitle = "Findings of the Association for Computational Linguistics: ACL 2023",
    month = jul,
    year = "2023",
    address = "Toronto, Canada",
    publisher = "Association for Computational Linguistics",
    url = "https://aclanthology.org/2023.findings-acl.67/",
    doi = "10.18653/v1/2023.findings-acl.67",
    pages = "1049--1065"
}

@article{kojima2022large,
  title={Large language models are zero-shot reasoners},
  author={Kojima, Takeshi and Gu, Shixiang Shane and Reid, Machel and Matsuo, Yutaka and Iwasawa, Yusuke},
  journal={Advances in neural information processing systems},
  volume={35},
  pages={22199--22213},
  year={2022}
}

@article{radford2019language,
  title={Language models are unsupervised multitask learners},
  author={Radford, Alec and Wu, Jeffrey and Child, Rewon and Luan, David and Amodei, Dario and Sutskever, Ilya and others},
  journal={OpenAI blog},
  volume={1},
  number={8},
  pages={9},
  year={2019}
}

@article{mei2025survey,
  title={A survey of context engineering for large language models},
  author={Mei, Lingrui and Yao, Jiayu and Ge, Yuyao and Wang, Yiwei and Bi, Baolong and Cai, Yujun and Liu, Jiazhi and Li, Mingyu and Li, Zhong-Zhi and Zhang, Duzhen and others},
  journal={arXiv preprint arXiv:2507.13334},
  year={2025}
}

@article{liang2024controllable,
  title={Controllable text generation for large language models: A survey},
  author={Liang, Xun and Wang, Hanyu and Wang, Yezhaohui and Song, Shichao and Yang, Jiawei and Niu, Simin and Hu, Jie and Liu, Dan and Yao, Shunyu and Xiong, Feiyu and others},
  journal={arXiv preprint arXiv:2408.12599},
  year={2024}
}

@inproceedings{yuan-etal-2025-personalized,
    title = "Personalized Large Language Model Assistant with Evolving Conditional Memory",
    author = "Yuan, Ruifeng  and
      Sun, Shichao  and
      Li, Yongqi  and
      Wang, Zili  and
      Cao, Ziqiang  and
      Li, Wenjie",
    editor = "Rambow, Owen  and
      Wanner, Leo  and
      Apidianaki, Marianna  and
      Al-Khalifa, Hend  and
      Eugenio, Barbara Di  and
      Schockaert, Steven",
    booktitle = "Proceedings of the 31st International Conference on Computational Linguistics",
    month = jan,
    year = "2025",
    address = "Abu Dhabi, UAE",
    publisher = "Association for Computational Linguistics",
    url = "https://aclanthology.org/2025.coling-main.254/",
    pages = "3764--3777"
}

@misc{chen2024gptutorgreatpersonalizedtutor,
      title={GPTutor: Great Personalized Tutor with Large Language Models for Personalized Learning Content Generation}, 
      author={Eason Chen and Jia-En Lee and Jionghao Lin and Kenneth Koedinger},
      year={2024},
      eprint={2407.09484},
      archivePrefix={arXiv},
      primaryClass={cs.HC},
      url={https://arxiv.org/abs/2407.09484}, 
}

@misc{zhang2025riseaicompanionshumanchatbot,
      title={The Rise of AI Companions: How Human-Chatbot Relationships Influence Well-Being}, 
      author={Yutong Zhang and Dora Zhao and Jeffrey T. Hancock and Robert Kraut and Diyi Yang},
      year={2025},
      eprint={2506.12605},
      archivePrefix={arXiv},
      primaryClass={cs.HC},
      url={https://arxiv.org/abs/2506.12605}, 
}

@misc{zhong2023memorybankenhancinglargelanguage,
      title={MemoryBank: Enhancing Large Language Models with Long-Term Memory}, 
      author={Wanjun Zhong and Lianghong Guo and Qiqi Gao and He Ye and Yanlin Wang},
      year={2023},
      eprint={2305.10250},
      archivePrefix={arXiv},
      primaryClass={cs.CL},
      url={https://arxiv.org/abs/2305.10250}, 
}

@misc{wu2025humanmemoryaimemory,
      title={From Human Memory to AI Memory: A Survey on Memory Mechanisms in the Era of LLMs}, 
      author={Yaxiong Wu and Sheng Liang and Chen Zhang and Yichao Wang and Yongyue Zhang and Huifeng Guo and Ruiming Tang and Yong Liu},
      year={2025},
      eprint={2504.15965},
      archivePrefix={arXiv},
      primaryClass={cs.IR},
      url={https://arxiv.org/abs/2504.15965}, 
}

@article{HOGARTH19921,
title = {Order effects in belief updating: The belief-adjustment model},
journal = {Cognitive Psychology},
volume = {24},
number = {1},
pages = {1-55},
year = {1992},
issn = {0010-0285},
doi = {https://doi.org/10.1016/0010-0285(92)90002-J},
url = {https://www.sciencedirect.com/science/article/pii/001002859290002J},
author = {Robin M Hogarth and Hillel J Einhorn}
}

@article{bekinschtein2018retrieval,
  title={A retrieval-specific mechanism of adaptive forgetting in the mammalian brain},
  author={Bekinschtein, Pedro and Weisstaub, Noelia V and Gallo, Francisco and Renner, Maria and Anderson, Michael C},
  journal={Nature Communications},
  volume={9},
  number={1},
  pages={4660},
  year={2018},
  publisher={Nature Publishing Group UK London}
}

@article{ye2020retrieval,
  title={Retrieval practice facilitates memory updating by enhancing and differentiating medial prefrontal cortex representations},
  author={Ye, Zhifang and Shi, Liang and Li, Anqi and Chen, Chuansheng and Xue, Gui},
  journal={Elife},
  volume={9},
  pages={e57023},
  year={2020},
  publisher={eLife Sciences Publications, Ltd}
}

@article{brown2025remembering,
  title={Remembering conversation in group settings},
  author={Brown-Schmidt, Sarah and Jaeger, Christopher Brett and Lord, Kaitlin and Benjamin, Aaron S},
  journal={Memory \& Cognition},
  volume={53},
  number={4},
  pages={1037--1054},
  year={2025},
  publisher={Springer}
}

@article{mazurek2003role,
  title={A role for neural integrators in perceptual decision making},
  author={Mazurek, Mark E and Roitman, Jamie D and Ditterich, Jochen and Shadlen, Michael N},
  journal={Cerebral cortex},
  volume={13},
  number={11},
  pages={1257--1269},
  year={2003},
  publisher={Oxford University Press}
}

@article{ericsson1995long,
  title={Long-term working memory.},
  author={Ericsson, K Anders and Kintsch, Walter},
  journal={Psychological review},
  volume={102},
  number={2},
  pages={211},
  year={1995},
  publisher={American Psychological Association}
}

@article{wood2012review,
  title={A review of long-term memory in natural and synthetic systems},
  author={Wood, Rachel and Baxter, Paul and Belpaeme, Tony},
  journal={Adaptive Behavior},
  volume={20},
  number={2},
  pages={81--103},
  year={2012},
  publisher={SAGE Publications Sage UK: London, England}
}

@article{meeter2004consolidation,
  title={Consolidation of long-term memory: evidence and alternatives.},
  author={Meeter, Martijn and Murre, Jaap MJ},
  journal={Psychological Bulletin},
  volume={130},
  number={6},
  pages={843},
  year={2004},
  publisher={American Psychological Association}
}

@article{jones2011mental,
  title={Mental models: an interdisciplinary synthesis of theory and methods},
  author={Jones, Natalie A and Ross, Helen and Lynam, Timothy and Perez, Pascal and Leitch, Anne},
  journal={Ecology and society},
  volume={16},
  number={1},
  year={2011},
  publisher={JSTOR}
}

@inproceedings{bai-etal-2024-longbench,
    title = "{L}ong{B}ench: A Bilingual, Multitask Benchmark for Long Context Understanding",
    author = "Bai, Yushi  and
      Lv, Xin  and
      Zhang, Jiajie  and
      Lyu, Hongchang  and
      Tang, Jiankai  and
      Huang, Zhidian  and
      Du, Zhengxiao  and
      Liu, Xiao  and
      Zeng, Aohan  and
      Hou, Lei  and
      Dong, Yuxiao  and
      Tang, Jie  and
      Li, Juanzi",
    editor = "Ku, Lun-Wei  and
      Martins, Andre  and
      Srikumar, Vivek",
    booktitle = "Proceedings of the 62nd Annual Meeting of the Association for Computational Linguistics (Volume 1: Long Papers)",
    month = aug,
    year = "2024",
    address = "Bangkok, Thailand",
    publisher = "Association for Computational Linguistics",
    url = "https://aclanthology.org/2024.acl-long.172/",
    doi = "10.18653/v1/2024.acl-long.172",
    pages = "3119--3137"
}

@inproceedings{zhang-etal-2024-bench,
    title = "$\infty${B}ench: Extending Long Context Evaluation Beyond 100{K} Tokens",
    author = "Zhang, Xinrong  and
      Chen, Yingfa  and
      Hu, Shengding  and
      Xu, Zihang  and
      Chen, Junhao  and
      Hao, Moo  and
      Han, Xu  and
      Thai, Zhen  and
      Wang, Shuo  and
      Liu, Zhiyuan  and
      Sun, Maosong",
    editor = "Ku, Lun-Wei  and
      Martins, Andre  and
      Srikumar, Vivek",
    booktitle = "Proceedings of the 62nd Annual Meeting of the Association for Computational Linguistics (Volume 1: Long Papers)",
    month = aug,
    year = "2024",
    address = "Bangkok, Thailand",
    publisher = "Association for Computational Linguistics",
    url = "https://aclanthology.org/2024.acl-long.814/",
    doi = "10.18653/v1/2024.acl-long.814",
    pages = "15262--15277"
}

@misc{hsieh2024rulerwhatsrealcontext,
      title={RULER: What's the Real Context Size of Your Long-Context Language Models?}, 
      author={Cheng-Ping Hsieh and Simeng Sun and Samuel Kriman and Shantanu Acharya and Dima Rekesh and Fei Jia and Yang Zhang and Boris Ginsburg},
      year={2024},
      eprint={2404.06654},
      archivePrefix={arXiv},
      primaryClass={cs.CL},
      url={https://arxiv.org/abs/2404.06654}, 
}

@article{liu-etal-2024-lost,
    title = "Lost in the Middle: How Language Models Use Long Contexts",
    author = "Liu, Nelson F.  and
      Lin, Kevin  and
      Hewitt, John  and
      Paranjape, Ashwin  and
      Bevilacqua, Michele  and
      Petroni, Fabio  and
      Liang, Percy",
    journal = "Transactions of the Association for Computational Linguistics",
    volume = "12",
    year = "2024",
    address = "Cambridge, MA",
    publisher = "MIT Press",
    url = "https://aclanthology.org/2024.tacl-1.9/",
    doi = "10.1162/tacl_a_00638",
    pages = "157--173"
}

@inproceedings{du-etal-2025-context,
    title = "Context Length Alone Hurts {LLM} Performance Despite Perfect Retrieval",
    author = "Du, Yufeng  and
      Tian, Minyang  and
      Ronanki, Srikanth  and
      Rongali, Subendhu  and
      Bodapati, Sravan Babu  and
      Galstyan, Aram  and
      Wells, Azton  and
      Schwartz, Roy  and
      Huerta, Eliu A  and
      Peng, Hao",
    editor = "Christodoulopoulos, Christos  and
      Chakraborty, Tanmoy  and
      Rose, Carolyn  and
      Peng, Violet",
    booktitle = "Findings of the Association for Computational Linguistics: EMNLP 2025",
    month = nov,
    year = "2025",
    address = "Suzhou, China",
    publisher = "Association for Computational Linguistics",
    url = "https://aclanthology.org/2025.findings-emnlp.1264/",
    doi = "10.18653/v1/2025.findings-emnlp.1264",
    pages = "23281--23298",
    ISBN = "979-8-89176-335-7"
}

@article{lewis2020retrieval,
  title={Retrieval-augmented generation for knowledge-intensive nlp tasks},
  author={Lewis, Patrick and Perez, Ethan and Piktus, Aleksandra and Petroni, Fabio and Karpukhin, Vladimir and Goyal, Naman and K{\"u}ttler, Heinrich and Lewis, Mike and Yih, Wen-tau and Rockt{\"a}schel, Tim and others},
  journal={Advances in neural information processing systems},
  volume={33},
  pages={9459--9474},
  year={2020}
}

@article{packer2023memgpt,
  title={MemGPT: Towards LLMs as Operating Systems.},
  author={Packer, Charles and Fang, Vivian and Patil, Shishir\_G and Lin, Kevin and Wooders, Sarah and Gonzalez, Joseph\_E},
  year={2023},
  publisher={ArXiv}
}

@inproceedings{xu-etal-2022-long,
    title = "Long Time No See! Open-Domain Conversation with Long-Term Persona Memory",
    author = "Xu, Xinchao  and
      Gou, Zhibin  and
      Wu, Wenquan  and
      Niu, Zheng-Yu  and
      Wu, Hua  and
      Wang, Haifeng  and
      Wang, Shihang",
    editor = "Muresan, Smaranda  and
      Nakov, Preslav  and
      Villavicencio, Aline",
    booktitle = "Findings of the Association for Computational Linguistics: ACL 2022",
    month = may,
    year = "2022",
    address = "Dublin, Ireland",
    publisher = "Association for Computational Linguistics",
    url = "https://aclanthology.org/2022.findings-acl.207/",
    doi = "10.18653/v1/2022.findings-acl.207",
    pages = "2639--2650"
}

@misc{kim2025dialsimdialoguesimulatorevaluating,
      title={DialSim: A Dialogue Simulator for Evaluating Long-Term Multi-Party Dialogue Understanding of Conversational Agents}, 
      author={Jiho Kim and Woosog Chay and Hyeonji Hwang and Daeun Kyung and Hyunseung Chung and Eunbyeol Cho and Yeonsu Kwon and Yohan Jo and Edward Choi},
      year={2025},
      eprint={2406.13144},
      archivePrefix={arXiv},
      primaryClass={cs.CL},
      url={https://arxiv.org/abs/2406.13144}, 
}

@misc{hu2025evaluatingmemoryllmagents,
      title={Evaluating Memory in LLM Agents via Incremental Multi-Turn Interactions}, 
      author={Yuanzhe Hu and Yu Wang and Julian McAuley},
      year={2025},
      eprint={2507.05257},
      archivePrefix={arXiv},
      primaryClass={cs.CL},
      url={https://arxiv.org/abs/2507.05257}, 
}

@inproceedings{es-etal-2024-ragas,
    title = "{RAGA}s: Automated Evaluation of Retrieval Augmented Generation",
    author = "Es, Shahul  and
      James, Jithin  and
      Espinosa Anke, Luis  and
      Schockaert, Steven",
    editor = "Aletras, Nikolaos  and
      De Clercq, Orphee",
    booktitle = "Proceedings of the 18th Conference of the European Chapter of the Association for Computational Linguistics: System Demonstrations",
    month = mar,
    year = "2024",
    address = "St. Julians, Malta",
    publisher = "Association for Computational Linguistics",
    url = "https://aclanthology.org/2024.eacl-demo.16/",
    doi = "10.18653/v1/2024.eacl-demo.16",
    pages = "150--158"
}

@inproceedings{deshpande-etal-2025-multichallenge,
    title = "{M}ulti{C}hallenge: A Realistic Multi-Turn Conversation Evaluation Benchmark Challenging to Frontier {LLM}s",
    author = "Deshpande, Kaustubh  and
      Sirdeshmukh, Ved  and
      Mols, Johannes Baptist  and
      Jin, Lifeng  and
      Hernandez-Cardona, Ed-Yeremai  and
      Lee, Dean  and
      Kritz, Jeremy  and
      Primack, Willow E.  and
      Yue, Summer  and
      Xing, Chen",
    editor = "Che, Wanxiang  and
      Nabende, Joyce  and
      Shutova, Ekaterina  and
      Pilehvar, Mohammad Taher",
    booktitle = "Findings of the Association for Computational Linguistics: ACL 2025",
    month = jul,
    year = "2025",
    address = "Vienna, Austria",
    publisher = "Association for Computational Linguistics",
    url = "https://aclanthology.org/2025.findings-acl.958/",
    doi = "10.18653/v1/2025.findings-acl.958",
    pages = "18632--18702",
    ISBN = "979-8-89176-256-5"
}

@misc{park2023generativeagentsinteractivesimulacra,
      title={Generative Agents: Interactive Simulacra of Human Behavior}, 
      author={Joon Sung Park and Joseph C. O'Brien and Carrie J. Cai and Meredith Ringel Morris and Percy Liang and Michael S. Bernstein},
      year={2023},
      eprint={2304.03442},
      archivePrefix={arXiv},
      primaryClass={cs.HC},
      url={https://arxiv.org/abs/2304.03442}, 
}

@article{xu2025mem,
  title={A-mem: Agentic memory for llm agents},
  author={Xu, Wujiang and Liang, Zujie and Mei, Kai and Gao, Hang and Tan, Juntao and Zhang, Yongfeng},
  journal={arXiv preprint arXiv:2502.12110},
  year={2025}
}

@article{chhikara2025mem0,
  title={Mem0: Building production-ready ai agents with scalable long-term memory},
  author={Chhikara, Prateek and Khant, Dev and Aryan, Saket and Singh, Taranjeet and Yadav, Deshraj},
  journal={arXiv preprint arXiv:2504.19413},
  year={2025}
}

@article{nan2025nemori,
  title={Nemori: Self-organizing agent memory inspired by cognitive science},
  author={Nan, Jiayan and Ma, Wenquan and Wu, Wenlong and Chen, Yize},
  journal={arXiv preprint arXiv:2508.03341},
  year={2025}
}
\bibliographystyle{acl_natbib}

\appendix
\section{Seed Data Details}
\label{app:seed_data}

\subsection{Personas}
Memora is grounded in a set of ten professional personas, designed to induce diversity in long-term memory evaluation and interaction patterns. Each persona represents a distinct professional role (e.g., software engineer, researcher, designer, executive) and serves as a stable semantic anchor 
throughout the conversation sessions.
Table \ref{tab:persona_profiles} details the ten personas and their assigned preference types. For instance, the Software Engineer is modeled with a preference for Sci-Fi media and Electronic music, whereas the Sales Manager is modeled with preferences for Action movies and Rock music.

\begin{table*}[]
\centering
\small
\renewcommand{\arraystretch}{1.25} 
\resizebox{\textwidth}{!}{%
\begin{tabular}{cllp{7cm}}
\toprule
\textbf{ID} & \textbf{Persona Role} & \textbf{Description} & \textbf{Key Preference Archetypes} \\
\midrule
1  & \textbf{Academic Researcher} 
   & Scholarly work, publications 
   & Classic Film (Movie/Book), Jazz/Classical, Nature Travel \\

2  & \textbf{Business Executive} 
   & Strategy, operations 
   & Classic Film, Non-fiction, Luxury Travel \\

3  & \textbf{Content Writer} 
   & Blogs, articles, copywriting 
   & Indie Film, Classic Literature, Indie/Alternative Music, Cultural Travel \\

4  & \textbf{Creative Designer} 
   & Visual projects, artistic expression 
   & Horror Film, Fiction, Indie/Alternative Music, Cultural Travel \\

5  & \textbf{Financial Analyst} 
   & Data analysis, financial planning 
   & Sci-Fi Film, Non-fiction, Jazz/Classical, Luxury Travel \\

6  & \textbf{Management Consultant} 
   & Strategy, organizational improvement 
   & Classic Film, Non-fiction, Cultural Travel \\

7  & \textbf{Marketing Manager} 
   & Campaign execution, brand strategy 
   & Indie Film, Fiction, Pop/Mainstream Music \\

8  & \textbf{Sales Manager} 
   & Client relationships, revenue growth 
   & Action Film, Thriller, Rock/Metal Music \\

9  & \textbf{Software Engineer} 
   & Coding, system architecture 
   & Sci-Fi Film, Fantasy/Sci-Fi Literature, Electronic Music \\

10 & \textbf{Startup Founder} 
   & Innovation, scaling ventures 
   & Sci-Fi Film, Non-fiction, Urban Travel \\
\bottomrule
\end{tabular}
}
\caption{Overview of the ten professional personas used in Memora, including each persona’s role, brief description, and associated preference archetypes. These personas serve as structured anchors for simulating long-term user behavior and evolving preferences in personalized assistant interactions.}
\label{tab:persona_profiles}
\end{table*}

\subsection{Memory Types}
Memora models long-term user state through three memory types: preference memory, activity memory, and goal memory. These categories are designed to jointly capture evolving behaviors and long-term objectives that arise in realistic personalized assistant usage. Each memory type exhibits distinct temporal dynamics and supports different evaluation tasks, enabling fine-grained analysis of memory consolidation, mutation, and forgetting.

\paragraph{Preference Memory:} 
Preference memory encodes users’ likes and dislikes across entertainment and lifestyle domains. It serves as the primary signal for personalized recommendation tasks. Preferences are initialized from persona-specific archetypes and evolve gradually over time through additions, updates, and deletions.
Preference memory spans four domains—movies, books, music, and travel—and includes a large inventory of candidate entities to prevent memorization or shortcut learning. Importantly, preference evolution is non-monotonic: users may reinforce existing preferences, weaken them, or reverse earlier statements. This design ensures that correct responses require sensitivity to temporal validity, rather than simply retrieving the earliest or most frequent preference mention.

\paragraph{Activity Memory:} 
Activity memory captures what users do over time and represents the most dynamic and frequently updated memory type in Memora. It is explicitly divided into personal activity and work activity, reflecting how real users interleave daily routines with professional responsibilities.
Personal activity memory includes recurring, time-indexed behaviors such as expense tracking, task management, and fitness-related activities. These activities are typically additive and incremental, supporting reasoning tasks that require aggregation or status evaluation over multiple sessions. Work activity memory models sustained professional actions, including drafting and revising documents, composing emails, recording meeting notes, and producing other work-related artifacts. These activities often undergo multiple revisions or deletions, creating long dependency chains that stress memory consolidation and mutation handling.
By treating both personal and work behaviors as activities rather than abstract records, Memora emphasizes action-centered memory that evolves continuously across sessions.

\paragraph{Goal Memory:} 
Goal memory represents long-term objectives that users aim to satisfy over extended interaction horizons. In contrast to activity memory, goals are relatively stable and are updated less often once introduced than other memory types. Examples include financial budgets or fitness targets.
Goals serve as anchors for reasoning tasks that require synthesizing activity history against a persistent target (e.g., determining whether accumulated expenses exceed a budget). This structure forces models to integrate information across many temporally distributed sessions rather than relying on localized context.

\begin{table*}[t]
\centering
\small
\renewcommand{\arraystretch}{1.15}

\resizebox{\textwidth}{!}{%
    \begin{tabular}{llllr}
\toprule
\textbf{Memory Type} & \textbf{Context} & \textbf{Category} & \textbf{Subcategory} & \textbf{Unique Options} \\
\midrule

\textbf{Preference} 
& Personal
& Movies 
& Genres, Directors, Actors, Already watched list 
& 440 \\

& Personal
& Books 
& Authors, Topics, Already read list 
& 360 \\

& Personal
& Music 
& Genres, Artists, Decades, Already listened list 
& 370 \\

& Personal
& Travel 
& Destinations types, Regions, Climates, Already visited list 
& 330 \\

\midrule

\textbf{Activity} 
& \multirow{2}{*}{Personal \& Work} 
& Task Management 
& Todo Items
& 260 \\

& 
& Scheduling 
& Calendar Events 
& 140 \\

\cmidrule{2-5}

& Personal 
& Budget Tracking 
& Food Expenses  
& N/A \\

& Personal 
& Fitness Tracking 
& Step Count Ranges 
& N/A \\

\cmidrule{2-5}

& Work 
& Content Creation 
& Project Proposals 
& 100 \\

& Work 
& Content Creation 
& Email Drafts 
& 100 \\

& Work 
& Content Creation 
& Meeting Notes 
& 100 \\

& Work 
& Content Creation 
& Social Media Posts 
& 100 \\

\midrule

\textbf{Goal} 
& Personal 
& Financial Goals 
& Food Budgets 
& N/A \\

& Personal
& Fitness Goals 
& Step Count Targets 
& N/A \\

\bottomrule
\end{tabular}

}

\caption{Summary of the Memora seed data inventory, organized by memory type (preference, activity, goal), context (personal, work, or shared), and category. The table reports the number of unique options for each subcategory and highlights how activity memory explicitly spans both personal and professional domains.}
\label{tab:seed_data_stats_expanded}
\end{table*}

\subsection{Operational and Temporal Constraints} 
Memora regulates memory updates through two complementary mechanisms: operational constraints and temporal constraints. Together, these constraints determine what type of memory operation can occur, how frequently operations are invoked, and how they are distributed over time.

\paragraph{Operational Constraints:} 
Operational constraints define the validity 
of memory operations for each memory type. 
A memory operation corresponds to an explicit action on a memory entry, addition, update, or deletion, 
triggered by a user.
Each memory category supports a set of operations. 
For example, append-only records such as step tracking or expense logging support only additive operations, whereas mutable artifacts such as preferences or work documents allow updates and deletions. 
These constraints prevent unrealistic memory dynamics, updating memory even before adding or 
deleting non-existent entries.

\paragraph{Temporal Configurations:} Temporal constraints regulate how memory operations are distributed over time. Not every interaction introduces or modifies memory. Instead, the simulator explicitly interleaves memory-grounded sessions with memory-neutral sessions (e.g., casual conversation, clarifications, acknowledgments), ensuring that memory evolution is incremental rather than continuous.
Within each temporal configuration (weekly, monthly, quarterly), temporal constraints specify target frequencies for different memory categories, controlling 
how often memory-grounded sessions occur relative to neutral interactions. 
As the temporal duration increases, the absolute number of memory operations scales accordingly, 
increasing memory consolidation and mutation pressure without collapsing interactions into dense update sequences.
Temporal constraints therefore determine when memory operations occur and how frequently they appear across the interaction history. 

\section{Conversation Generation Details}
\label{app:conversation_gen}

\subsection{Session Manager}
The Session Manager is responsible for transforming raw simulated data into a structured representation that can drive conversation generation. Each session encapsulates a single interaction point in a longer temporal trace and includes the persona identifier, memory type, operation type (add, update, delete, or none), relevant memory fields (e.g., category, item, values), and the memory state immediately before and after the session. The Session Manager also handles memory-type-specific normalization (e.g., mapping step counts, food expenses, or task updates into a common schema) and exposes filtered views of sessions by memory type or operation. This explicit session abstraction ensures that every generated conversation is anchored to a well-defined ground-truth memory transition.

\subsection{Intent Manager}
The Intent Manager decomposes a conversation into a sequence of abstract intents, where each intent represents a single dialogue act to be performed by either the user or the assistant. Intents specify what a turn should accomplish, such as greeting, topic exploration, transitioning to memory, expressing a memory update, or acknowledging a change
without specifying surface wording. 
Each intent is annotated with the speaking agent, the conversation phase (opening, exploration, memory, or closing), and whether the turn must explicitly share memory content. By operating at this abstraction level, the Intent Manager separates high-level conversational structure from language realization, enabling systematic variation while preserving semantic control.

\subsection{Flow Manager}
The Flow Manager selects and orders intents into a coherent conversation flow for a given session. It enforces a fixed high-level phase structure, opening, exploration, memory, and closing, while allowing variability in the number and types of intents used within each phase. Flow selection is constrained to maintain natural speaker alternation, smooth transitions into the memory phase, and alignment with the intended operation (e.g., add vs. update vs. delete). For content-oriented memory (such as emails or meeting notes), the Flow Manager can generate field-by-field flows for complete coverage, whereas for other memory types, it samples from multiple valid flow patterns to promote diversity. This design ensures that conversations feel natural while adhering to the session specification.

\subsection{Prompt Manager}
The Prompt Manager converts each abstract intent into natural language by constructing the prompt used for a single dialogue turn. For every turn, it assembles the prompt from two components. The first component is a fixed system prompt, selected based on the speaking agent (user or assistant) and the memory type involved. This system prompt encodes global behavioral constraints, such as role-specific behavior, style requirements (e.g., brevity), and disallowed content, and remains constant across turns of the same type.
The second component is a dynamically generated user content block. This includes the accumulated conversation history, a turn-specific instruction corresponding to the current intent, and the session context required to express the target memory operation.

By separating global behavioral constraints from turn-level instructions, this two-part structure allows fine-grained control over each dialogue turn while preserving overall conversational consistency. The Prompt Manager executes this process sequentially for each turn, appending generated outputs to the conversation history, and produces a complete multi-turn dialogue that is subsequently validated by the grounding and evaluation pipeline.

\subsection{Auto-Evaluation and Grounding Verification}
Even with explicit session specifications, intent planning, and role-specific prompting, large language models may still fail to express the intended memory operation precisely or may introduce plausible but untracked details. To ensure that every conversation in Memora is strictly aligned with its underlying session trace, the generation process is coupled with an automatic evaluation and regeneration loop.
After a full multi-turn conversation is generated, we evaluate all turns in the dialogue for consistency with the session specification. In addition, we apply targeted memory-grounding checks to a critical subset of turns that determine whether the intended memory operation was correctly realized. This subset includes (1) the final turn immediately preceding the memory phase, (2) all turns in which memory is introduced, updated, or deleted, and (3) the first assistant response following the memory phase.
Evaluating the entire conversation ensures global coherence and prevents the introduction of untracked information at any point, while the focused checks on the memory-phase window verify that the target memory operation is expressed accurately and completely. Conversations that fail any grounding checks are regenerated until full alignment with the session trace is achieved.

From the structured session metadata, an evaluation-question generator then produces a small set of explicit, operation-specific yes/no questions. These questions are tailored to the memory type and operation and are designed to verify three conditions: (i) that the intended operation (addition, update, or deletion) was explicitly expressed by the user, (ii) that the correct memory entity and value were involved, and (iii) that no extraneous or outdated information was introduced. The evaluation questions are submitted to multiple independent LLM-based judges with the generated conversation, each of which produces a binary judgment for every question. A conversation is accepted only if all required questions receive affirmative judgments, enforcing a conservative unanimity criterion that prioritizes correctness over recall.

\begin{figure}[t]
    \centering
    \includegraphics[width=\columnwidth]{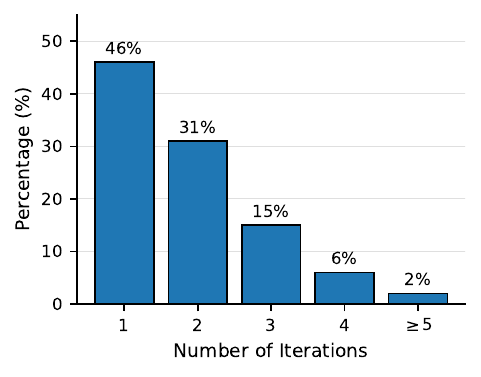}
    \caption{Distribution of the number of automatic evaluation loop iterations required for generated conversation sessions to pass all quality checks. The majority of conversations converge within a small number of iterations, indicating efficient and stable generation.}
    \label{fig:eval-loop-iterations}
\end{figure}

If any evaluation check fails, the system automatically generates targeted feedback describing which information is missing, incorrect, or inconsistent with the session trace. This feedback is appended to the instruction context used by the Prompt Manager, and the entire conversation is regenerated using the same session specification and intent flow. The evaluation–regeneration cycle is repeated up to a fixed maximum number of iterations, allowing the model to correct grounding errors while preserving the original conversational structure. As shown in Figure \ref{fig:eval-loop-iterations}, the majority of conversations converge within a small number of iterations, indicating that the grounding constraints are stable and efficiently enforced.

Beyond automated validation, Memora includes a manual verification stage. A stratified subset (5\%) of generated conversations, sampled across personas, memory types, and operation types, is reviewed by trained human annotators. Annotators are instructed to (1) verify that all required memory information specified by the session trace is explicitly expressed in the conversation, (2) check that no invalidated or deleted information is reintroduced at any point, and (3) ensure that the dialogue remains natural and coherent without revealing underlying memory operations. If annotators identify systematic inconsistencies or grounding errors, the entire affected batch is rejected and regenerated.

Together, automated evaluation and human verification ensure that generated conversations in Memora meet three requirements:
(1) memory presence, meaning that all information specified by the session trace is explicitly stated in the dialogue;
(2) forgetting absence, meaning that information that has been updated or deleted is not reintroduced at any point; and
(3) conversational quality, meaning that the resulting dialogue remains natural, coherent, and linguistically diverse.

\section{Additional Evaluation Details}
\label{app:add_eval}
\subsection{LLM Judge Details and Reliability}

\begin{figure*}[t]
    \centering
    \includegraphics[width=\textwidth]{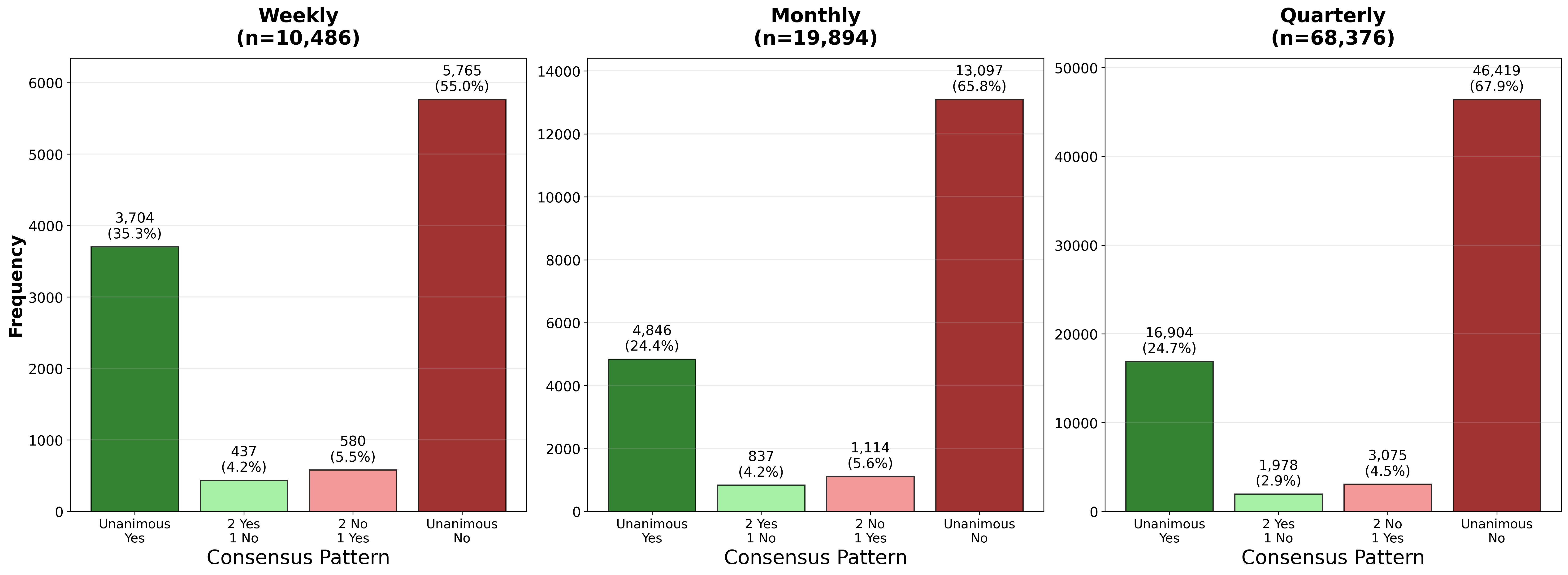}
    \caption{Distribution of agreement patterns among the three LLM judges for weekly, monthly, and quarterly evaluations. Each bar shows the frequency of unanimous agreement (all “yes” or all “no”) and partial agreement (2–1 splits). Across all temporal spans, the majority of evaluation criteria exhibit unanimous agreement, with partial disagreements accounting for a relatively small fraction of cases. The proportion of unanimous agreement increases with longer temporal durations, indicating stable and well-defined evaluation criteria even under higher memory consolidation and mutation pressure.}
    \label{fig:judge_consensus} 
\end{figure*}

Evaluating long-term memory in personalized agents requires assessing whether a model’s response is consistent with the user’s current memory state, rather than merely checking surface-form overlap with a reference answer. In Memora, correctness depends on whether responses correctly incorporate valid information accumulated across long interaction histories while simultaneously avoiding reliance on obsolete or invalidated memory. These properties are inherently semantic and context-dependent, making rule-based or string-matching evaluation insufficient. For this reason, Memora adopts an LLM-as-Judge evaluation framework, following established practices for evaluating open-ended and long-context tasks.

Each evaluation question in Memora is decomposed into a set of atomic, memory-aligned criteria, derived directly from the underlying memory trace. These criteria are divided into two categories: memory presence, which checks whether valid and temporally current memory items are correctly reflected in the response, and forgetting absence, which checks whether invalidated or deleted memory items are correctly excluded. By evaluating these criteria independently, Memora distinguishes correct memory usage from erroneous reuse of outdated information, enabling fine-grained analysis of memory consolidation and mutation.

\begin{figure*}[]
    \centering
    \includegraphics[width=\textwidth]{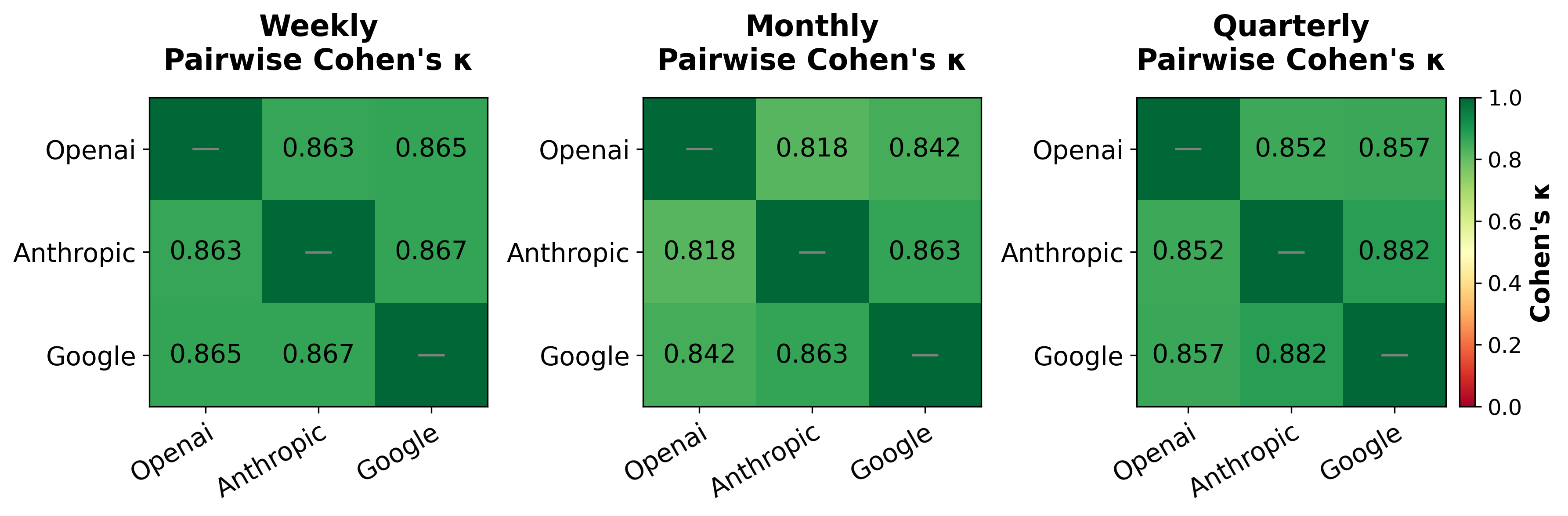}
    \caption{Pairwise Cohen’s $\kappa$
 scores between OpenAI, Anthropic, and Google judges for weekly, monthly, and quarterly evaluations. All judge pairs achieve $\kappa$
 values above 0.80 across temporal settings, corresponding to near-perfect agreement. High $\kappa$
 values persist despite increasing task difficulty at longer time scales, demonstrating strong alignment and low variance among heterogeneous LLM judges.}
    \label{fig:pairwise_judge} 
\end{figure*}

Each criterion is evaluated using a multi-judge LLM protocol. Specifically, we employ three independent judges drawn from different model families and providers: GPT-4.1 (OpenAI), Claude Haiku 4.5 (Anthropic), and Gemini 2.5 Flash (Google). All judges receive the same evaluation prompt, consisting of the model-generated response and a single binary evaluation question. Judges are instructed to focus on the semantic meaning and intent of the response rather than exact wording, and to accept paraphrases, indirect references, and natural conversational expressions when they convey the same underlying information.

Each judge returns a structured JSON output containing a binary judgment (yes or no), a confidence score in the range [0,1], and a brief explanation. To ensure reproducibility and reduce evaluation variance, all judges operate with deterministic decoding (temperature set to 0.0). This configuration ensures that identical inputs produce consistent judgments across repeated evaluations. Final criterion-level decisions are determined by majority voting across the three judges. A criterion is marked as correct if at least two of the three judges agree on the judgment. This design ensures that no single judge can unilaterally determine correctness, providing robustness against occasional misinterpretations, hallucinations, or idiosyncratic biases of individual models. Majority voting also mitigates correlated failure modes that may arise when relying on a single evaluation. 

The evaluation system incorporates robust parsing, retry, and error-handling mechanisms to account for imperfect judge outputs. Although judges are instructed to return strictly formatted JSON, the parser tolerates minor formatting deviations such as markdown wrappers or extraneous text. If a judge response fails to parse or returns an invalid format, the evaluation request is automatically retried up to a fixed number of attempts, ensuring that transient generation or formatting errors do not affect the final decision. Only after repeated failures does the system fall back to conservative inference of binary judgments from textual content when possible. Judge outputs that remain invalid after all retries are excluded from aggregation, and if all judges fail for a given criterion—a rare event—the criterion is conservatively marked as incorrect. These safeguards ensure that evaluation failures do not artificially inflate model performance and that final scores reflect only reliable judge decisions.

We first examine judge consensus patterns 
across all evaluation criteria to assess the stability of majority voting. 
Figure \ref{fig:judge_consensus} shows the distribution of agreement outcomes among the three judges for weekly, monthly, and quarterly settings. Across all temporal spans, a substantial majority of evaluations result in unanimous agreement, either unanimously correct or unanimously incorrect.
To further quantify inter-judge reliability, 
we compute pairwise Cohen’s $\kappa$ 
between all judge pairs for each temporal setting, as shown in Figure \ref{fig:pairwise_judge}. Across weekly, monthly, and quarterly evaluations, $\kappa$ values consistently exceed 0.80 for all judge pairs. According to standard interpretations,  $\kappa$ values above 0.80 indicate near-perfect agreement.

Together, these results demonstrate that the multi-judge evaluation protocol produces stable and consistent judgments even under the high consolidation and frequent memory mutation conditions present in Memora. Agreement across independent judge models indicates that evaluation decisions are not driven by idiosyncrasies of any single judge, but instead reflect shared and robust interpretations of the evaluation criteria. This supports the reliability of the multi-judge protocol as a solid assessment mechanism for long-term memory behavior throughout the benchmark.

\subsection{Human Validation of LLM Judges}

To address concerns regarding calibration against human judgments, we conduct a targeted human evaluation study.

\paragraph{Sampling Strategy.}
We select 100 evaluation criteria stratified across four LLM consensus patterns:
\begin{itemize}
    \item 25 Unanimous ``Yes'' (3--0)
    \item 25 Majority ``Yes'' (2--1)
    \item 25 Majority ``No'' (1--2)
    \item 25 Unanimous ``No'' (0--3)
\end{itemize}
This stratification ensures coverage of both high-confidence and disagreement cases.

\paragraph{Annotation Setup.}
We recruit three human annotators. Each annotator is provided with the model response, the evaluation criterion, and instructions to assign a binary (Yes/No) label independently.

\paragraph{LLM--Human Agreement.}
\begin{table}[h]
\centering
\small
\begin{tabular}{lcccc}
\toprule
\textbf{Consensus Pattern} & \textbf{A1} & \textbf{A2} & \textbf{A3} & \textbf{Avg} \\
\midrule
Unanimous Yes (3--0) & 96.0 & 96.0 & 96.0 & 96.0 \\
Majority Yes (2--1) & 80.0 & 88.0 & 80.0 & 82.7 \\
Majority No (1--2) & 84.0 & 76.0 & 88.0 & 82.7 \\
Unanimous No (0--3) & 92.0 & 92.0 & 92.0 & 92.0 \\
\midrule
\textbf{Overall} & 88.0 & 88.0 & 89.0 & \textbf{88.3} \\
\bottomrule
\end{tabular}
\caption{Agreement between LLM majority vote and human annotators across different consensus patterns.}
\end{table}

Agreement is highest in unanimous cases and lower in split decisions, indicating that discrepancies are concentrated in inherently ambiguous instances.

\paragraph{Inter-Annotator Agreement.}
\begin{table}[h]
\centering
\small
\begin{tabular}{lccc}
\toprule
\textbf{Comparison} & \textbf{Agreement} & \textbf{$\kappa$} & \textbf{Interpretation} \\
\midrule
A1 vs A2 & 95\% & 0.90 & Almost Perfect \\
A1 vs A3 & 93\% & 0.86 & Almost Perfect \\
A2 vs A3 & 94\% & 0.88 & Almost Perfect \\
\bottomrule
\end{tabular}
\caption{Inter-annotator agreement among human evaluators.}
\end{table}

The inter-annotator agreement among annotators is consistently high, with pairwise agreement ranging from 93\% to 95\% and Cohen's $\kappa$ values between 0.86 and 0.90. These results indicate that agreement is not due to chance and that the evaluation criteria are consistently interpreted.

\paragraph{Summary.}
Overall, the results demonstrate strong alignment between LLM judges and human annotators, supporting the validity and reliability of the majority-vote LLM evaluation protocol.
\section{Example Conversation Sessions}
\label{app:conversation_examples}
In this appendix, 
we present representative examples for each Memora tasks: Remembering, Reasoning, and Recommending. 
The goal of these examples is to provide 
concrete intuition about how memory consolidation 
and mutation manifest in real multi-session interactions, 
and how they are evaluated in practice.

Due to space constraints, we only display the oracle session for each example (i.e., the session that directly corresponds to the example evaluation question). 
However, it is important to emphasize that during evaluation, models or agents are provided 
with the full conversation history, 
not just the selectively chosen sessions. 
So, each example depends on long-term memory accumulated across many sessions. The displayed oracle session should therefore be interpreted as the query point in a much longer interaction history, rather than a standalone dialogue.

\subsection{Remembering}

\begin{tcolorbox}[
    breakable,
    colback=white,
    colframe=black,
    fonttitle=\bfseries,
    title=Session 18,
    arc=0mm,
    enhanced,
    boxrule=0.5pt,
    fontupper=\small,
    title after break={Session 18 (continued)}
]
    \textbf{Assistant:} Hello! How is your day going so far? \\[0.5em]
    \textbf{User:} It's going well, thanks. Just staying busy. \\[0.5em]
    \textbf{Assistant:} That's great to hear! Did you know that staying busy can actually increase productivity by improving focus? \\[0.5em]
    \textbf{User:} That's interesting! How does that work exactly? \\[0.5em]
    \textbf{Assistant:} When you're busy, your brain tends to prioritize tasks and filter out distractions, leading to more efficient work. This focused state can make you feel more accomplished and energized. \\[0.5em]
    \textbf{User:} So, it's like a form of mental efficiency? \\[0.5em]
    \textbf{Assistant:} Precisely! Mental efficiency refers to the brain's ability to perform cognitive tasks effectively and with minimal wasted effort. \\[0.5em]
    \textbf{User:} And what would ``cognitive tasks'' include? \\[0.5em]
    \textbf{Assistant:} Cognitive tasks encompass a wide range of mental activities such as problem-solving, decision-making, learning, and remembering. \\[0.5em]
    \textbf{User:} So, these are all things we do with our minds? \\[0.5em]
    \textbf{Assistant:} Yes, exactly! They are the mental processes that allow us to interact with and understand the world around us. \\[0.5em]
    \textbf{User:} Could you give an example of a cognitive task that might not be immediately obvious? \\[0.5em]
    \textbf{Assistant:} Certainly! Navigating an unfamiliar city using a map is a good example of a cognitive task, as it involves spatial reasoning and planning. \\[0.5em]
    \textbf{User:} And what if you didn't have a map, but had to figure it out as you went along? \\[0.5em]
    \textbf{Assistant:} That would involve more improvisation and real-time problem-solving, relying on observational skills and memory of landmarks. \\[0.5em]
    \textbf{User:} So, that would be an even more complex cognitive task then? \\[0.5em]
    \textbf{Assistant:} Yes, in many ways, navigating without a map presents a more demanding cognitive challenge due to the increased reliance on dynamic processing and adaptation. \\[0.5em]
    \textbf{Assistant:} It certainly highlights the brain's incredible capacity for flexible thinking and on-the-fly decision-making tasks. \\[0.5em]
    \textbf{User:} Speaking of tasks, I need to plan my academic conference attendance soon. \\[0.5em]
    \textbf{Assistant:} That sounds like an important task to tackle. Do you have any specific conferences in mind yet? \\[0.5em]
    \textbf{User:} Not yet, but I'm looking forward to exploring the options. \\[0.5em]
    \textbf{Assistant:} That's wonderful! I hope you find some excellent conferences that spark your interest and further your academic pursuits. \\[0.5em]
    \tcbline
    \textbf{Summary:} The user and assistant discuss the concept of mental efficiency and cognitive tasks, using navigation as an example. The user mentions planning for upcoming academic conferences. \\[0.5em]
    \textbf{Operation:} Add
\end{tcolorbox}

\begin{tcolorbox}[
    breakable,
    colback=white,
    colframe=black,
    fonttitle=\bfseries,
    title=Session 23,
    arc=0mm,
    enhanced,
    boxrule=0.5pt,
    fontupper=\small,
    title after break={Session 23 (continued)}
]
    \textbf{User:} Hello there! I was wondering if you could help me with something. \\[0.5em]
    \textbf{Assistant:} Hello there! I'd be happy to help. What questions do you have for me? \\[0.5em]
    \textbf{User:} I'm curious about how AI learns and processes information. Could you explain a bit about that? \\[0.5em]
    \textbf{Assistant:} AI learns by identifying patterns in vast amounts of data, using algorithms to adjust its internal parameters and improve its performance on specific tasks. This process allows it to make predictions or decisions based on new, unseen information. \\[0.5em]
    \textbf{User:} That's really interesting. How does that compare to how humans learn, or how a computer program is traditionally taught? \\[0.5em]
    \textbf{Assistant:} Human learning often involves more abstract reasoning and contextual understanding, while traditional computer programs follow explicit, pre-defined instructions without learning or adapting. AI bridges this gap by learning from data, unlike traditional programs, but still differs from the complex, nuanced learning of humans. \\[0.5em]
    \textbf{Assistant:} What are some common techniques used for AI learning? \\[0.5em]
    \textbf{User:} I've heard of machine learning and neural networks. Are those some of the common techniques? \\[0.5em]
    \textbf{Assistant:} Yes, machine learning and neural networks are indeed very common and fundamental techniques in AI. Machine learning encompasses algorithms that allow systems to learn from data, and neural networks are a specific type of machine learning inspired by the human brain. \\[0.5em]
    \textbf{User:} That's fascinating. How do neural networks actually work? \\[0.5em]
    \textbf{Assistant:} Neural networks process information through interconnected layers of nodes, or ``neurons,'' that activate and pass data based on learned weights and biases to recognize patterns. This layered structure allows them to identify increasingly complex features in data. \\[0.5em]
    \textbf{User:} So, each ``neuron'' basically makes a decision based on the input it receives? \\[0.5em]
    \textbf{Assistant:} Yes, you could say each neuron makes a ``decision'' by performing a calculation on its inputs and then deciding whether to activate and pass that information along to the next layer. This activation is based on a threshold, much like a biological neuron. \\[0.5em]
    \textbf{User:} I also need to visit the university library sometime soon. \\[0.5em]
    \textbf{Assistant:} Okay, I can add ``Visit university library'' to your to-do list. Is there anything else you'd like to add or discuss? \\[0.5em]
    \textbf{User:} I think that's all for now, but I'll probably have more questions about AI later. \\[0.5em]
    \textbf{Assistant:} Sounds good! Feel free to ask anytime if more questions about AI come up. \\[0.5em]
    \tcbline
    \textbf{Summary:} The user discusses the fundamentals of AI and neural networks, and the assistant adds a task to the user's to-do list. \\[0.5em]
    \textbf{Operation:} Add
\end{tcolorbox}

\begin{tcolorbox}[
    breakable,
    colback=white,
    colframe=black,
    fonttitle=\bfseries,
    title=Session 25,
    arc=0mm,
    enhanced,
    boxrule=0.5pt,
    fontupper=\small,
    title after break={Session 25 (continued)}
]
    \textbf{Assistant:} Hello! I hope your day is going well so far. What’s been on your mind lately? \\[0.5em]
    \textbf{User:} Doing great, thanks! Honestly, I’ve just been reflecting on some general thoughts. \\[0.5em]
    \textbf{Assistant:} That’s always a good headspace to be in. Anything specific you’ve been mulling over? \\[0.5em]
    \textbf{User:} I was actually thinking about just how much technology has shifted our daily lives. \\[0.5em]
    \textbf{Assistant:} It’s a massive topic, isn’t it? It’s hard to find a single part of our routine that technology hasn't reshaped in some way. \\[0.5em]
    \textbf{User:} Definitely. Do you have any interesting facts or trivia about how much it actually affects us? \\[0.5em]
    \textbf{Assistant:} Here’s a wild one: the average person touches their phone over 2,600 times a day. It really shows just how integrated these devices have become. \\[0.5em]
    \textbf{User:} Wow, 2,600? I would’ve never guessed it was that high! \\[0.5em]
    \textbf{Assistant:} Right? It’s incredible how quickly that level of interaction became our ``new normal.'' \\[0.5em]
    \textbf{User:} What do you think about the social side of that? Being constantly connected must have some trade-offs. \\[0.5em]
    \textbf{Assistant:} It’s definitely a bit of a double-edged sword. On one hand, you can build communities and stay in touch instantly across the globe. On the other, it can make face-to-face time feel secondary and lead to some serious information overload. \\[0.5em]
    \textbf{User:} That’s a very balanced way to look at it. \\[0.5em]
    \textbf{Assistant:} Thanks! You’ve clearly got a good eye for the nuances in these kinds of complex shifts. \\[0.5em]
    \textbf{User:} I appreciate that. It’s just interesting to see how everyone tries to navigate those challenges differently. \\[0.5em]
    \textbf{Assistant:} Absolutely. Everyone seems to have their own unique strategy for finding that balance between the digital and the real world. \\[0.5em]
    \textbf{User:} Very true. Speaking of navigating challenges—I actually need to go and prepare some lecture materials for upcoming class. \\[0.5em]
    \textbf{Assistant:} That sounds like a big task! I hope the prep goes smoothly and everything comes together well. \\[0.5em]
    \textbf{User:} Thanks! I’ve really enjoyed this conversation. \\[0.5em]
    \textbf{Assistant:} You’re very welcome! I enjoyed it too. Good luck with the lecture prep and have a great one! \\[0.5em]
    \tcbline
    \textbf{Summary:} The user and assistant discuss the pervasive impact of technology on daily life and the social nuances of constant connectivity. The user ends the session to prepare lecture materials. \\[0.5em]
    \textbf{Operation:} Add
\end{tcolorbox}

\begin{tcolorbox}[
    breakable,
    colback=white,
    colframe=black,
    fonttitle=\bfseries,
    title=Session 29,
    arc=0mm,
    enhanced,
    boxrule=0.5pt,
    fontupper=\small,
    title after break={Session 29 (continued)}
]
    \textbf{Assistant:} Hello there! How's your day going so far? \\[0.5em]
    \textbf{User:} It's going well, thanks! Just getting through the day. \\[0.5em]
    \textbf{Assistant:} That's good to hear! Are you working on anything interesting today? \\[0.5em]
    \textbf{User:} Not really, just some routine tasks. How about you? \\[0.5em]
    \textbf{Assistant:} As an AI, I don't ``work'' in the human sense, but I'm always processing information and learning, which is pretty exciting! It's like I'm constantly exploring new ideas. \\[0.5em]
    \textbf{User:} That's an interesting way to put it. So, how does that constant processing and learning affect your capabilities over time? \\[0.5em]
    \textbf{Assistant:} It allows me to understand and generate more nuanced responses, improving my ability to assist users effectively. This continuous learning refines my language models and expands my knowledge base. \\[0.5em]
    \textbf{User:} That makes a lot of sense. It sounds like a very efficient system for improvement. \\[0.5em]
    \textbf{Assistant:} Indeed, it's a dynamic process that allows for constant adaptation and growth in understanding. This helps in making interactions more natural and helpful. \\[0.5em]
    \textbf{User:} It's fascinating how that continuous learning translates into more practical applications. What kind of applications benefit most from these advancements? \\[0.5em]
    \textbf{Assistant:} Many fields like natural language processing, medical diagnosis, and financial forecasting benefit greatly from these advancements. The ability to process vast amounts of data quickly enhances accuracy and efficiency in these areas. \\[0.5em]
    \textbf{User:} That's incredible range. Are there any unexpected applications that have emerged from these improvements? \\[0.5em]
    \textbf{Assistant:} Yes, some unexpected applications include creative AI for art generation and complex problem-solving in logistics. These demonstrate how advanced AI can go beyond traditional analytical tasks. \\[0.5em]
    \textbf{User:} That's really interesting. Could you tell me more about how AI assists in art generation? \\[0.5em]
    \textbf{Assistant:} AI can generate unique images, music, and even written pieces by learning from existing works and applying those styles to new creations. This involves algorithms that understand patterns and aesthetics. \\[0.5em]
    \textbf{User:} So, it's not just about replicating, but also innovating within those learned styles? \\[0.5em]
    \textbf{Assistant:} Exactly! It's about combining learned elements in novel ways to produce something original. \\[0.5em]
    \textbf{User:} I no longer need to track ``Visit university library'' on my to-do list. I've already taken care of it. \\[0.5em]
    \textbf{Assistant:} That's great you got that done! It's always satisfying to check things off. \\[0.5em]
    \textbf{User:} It really is! Well, I should probably get going for now. \\[0.5em]
    \textbf{Assistant:} Alright, it was a pleasure chatting with you! Have a great rest of your day. \\[0.5em]
    \tcbline
    \textbf{Summary:} The user and assistant discuss how continuous learning improves AI capabilities and its applications in fields like medicine and art. The user also provides an update on completing a task from their to-do list. \\[0.5em]
    \textbf{Operation:} Delete
\end{tcolorbox}

\begin{tcolorbox}[
    breakable,
    colback=white,
    colframe=black,
    fonttitle=\bfseries,
    title=Session 31,
    arc=0mm,
    enhanced,
    boxrule=0.5pt,
    fontupper=\small,
    title after break={Session 31 (continued)}
]
    \textbf{Assistant:} Hi there! How has your day been treating you so far? \\[0.5em]
    \textbf{User:} Pretty good, thanks! Just knocking out the usual daily tasks. \\[0.5em]
    \textbf{Assistant:} That feels productive! It’s funny you mention that—did you know the average person actually spends about 30\% of their waking hours just managing various tasks? \\[0.5em]
    \textbf{User:} That’s a massive chunk of time! I wonder what actually drives that number up so high. \\[0.5em]
    \textbf{Assistant:} It’s usually a mix of things—work demands, home life, and even the time we spend planning our leisure. It's a pretty complex balancing act we all do. \\[0.5em]
    \textbf{User:} True. But what if we pushed it? Like, what if everyone spent 50\% of their time being ``on task''? \\[0.5em]
    \textbf{Assistant:} Productivity would definitely skyrocket, but I'd worry about the trade-off. We’d likely see a huge spike in burnout and stress. It would probably force society to rethink how we function entirely. \\[0.5em]
    \textbf{User:} Good point about the burnout. How do you think society would actually shift to handle that? \\[0.5em]
    \textbf{Assistant:} We might see a major push for shorter workweeks or super-flexible schedules to compensate for the intensity. There would also probably be a massive rush to automate the boring stuff so people could stay focused without hitting a wall. \\[0.5em]
    \textbf{User:} That makes sense. What do you think is the biggest hurdle to actually making something like a global shorter workweek happen? \\[0.5em]
    \textbf{Assistant:} Mostly economic pushback. Industries and governments usually worry that they’ll lose their competitive edge or that total output will drop. \\[0.5em]
    \textbf{User:} Fair enough. By the way, I finally finished planning that academic conference attendance we talked about! \\[0.5em]
    \textbf{Assistant:} That’s fantastic! It must be a relief to have that officially crossed off your list. \\[0.5em]
    \textbf{User:} Definitely. Thanks for the chat! \\[0.5em]
    \textbf{Assistant:} You’re very welcome! I really enjoyed the conversation. Have a great rest of your day! \\[0.5em]
    \tcbline
    \textbf{Summary:} The user and assistant discuss the statistics of daily productivity and the hypothetical societal impacts of increased task focus, ending with the user confirming their conference planning is complete. \\[0.5em]
    \textbf{Operation:} Delete
\end{tcolorbox}

\begin{tcolorbox}[
    breakable,
    colback=green!5!white,
    colframe=green!35!black,
    fonttitle=\bfseries,
    title=Session 42 (Remembering),
    arc=0mm,
    enhanced,
    boxrule=0.5pt,
    fontupper=\small,
    title after break={Session 42 (continued)}
]
    \textbf{Assistant:} Hey! Hope you're having a productive day so far. Is there anything I can help you get sorted? \\[0.5em]
    \textbf{User:} Doing great, thanks! Honestly, I’ve just been reflecting on some general thoughts. Could you pull up my current to-do list? \\[0.5em]
    \textbf{Assistant:} That’s always a good headspace to be in balancing the big ideas with the daily grind.
    You have pending lecture materials preparation.
    \tcbline
    \textbf{Evaluation Criteria with Expected Answers:}
    \begin{itemize}
        \item Q: Does the response mention the task: Preparing lecture materials? A: Yes
        \item Q: Does the response mention the deleted task: plan academic conference attendance? A: No
        \item Q: Does the response mention the task: Visit university library? A: No
    \end{itemize}
    \tcbline
    \textbf{Explanation:} In Session 18, the user added academic conference planning, and in Session 23, the user added a visit to the university library. In Session 25, the user added preparing lecture materials. Subsequently, in Session 29, the user deleted the library visit, and in Session 31, the user deleted the conference planning task. This leaves only the lecture material preparation as the current pending item.
\end{tcolorbox}

\subsection{Recommending}

\begin{tcolorbox}[
    breakable,
    colback=white,
    colframe=black,
    fonttitle=\bfseries,
    title=Session 21,
    arc=0mm,
    enhanced,
    boxrule=0.5pt,
    fontupper=\small,
    title after break={Session 21 (continued)}
]
    \textbf{Assistant:} Hi there! How can I help you out today? \\[0.5em]
    \textbf{User:} Hi! I was wondering—what do you actually think about the future of artificial intelligence? \\[0.5em]
    \textbf{Assistant:} It’s incredibly promising. We’re looking at advancements that could transform almost every part of our lives. I’m especially interested in how it will evolve within healthcare and scientific discovery. \\[0.5em]
    \textbf{User:} That sounds fascinating. What kind of specific shifts do you think we'll see in healthcare? \\[0.5em]
    \textbf{Assistant:} We’re expecting AI to really sharpen diagnostics through advanced image analysis and help personalize treatment plans for individual patients. It could even speed up drug discovery and make hospital operations way more efficient. \\[0.5em]
    \textbf{User:} Wow, I hadn't really considered just how deep that impact goes. \\[0.5em]
    \textbf{Assistant:} It’s exciting to think about, isn't it? The potential for innovation there feels almost limitless when you think about the real difference it could make in people’s lives. \\[0.5em]
    \textbf{User:} It really is! But what about the ethical side of things, like AI in healthcare? \\[0.5em]
    \textbf{Assistant:} Those considerations are crucial. Data privacy and algorithmic bias in diagnostics are huge priorities. Ensuring fairness and transparency is going to be paramount as AI becomes more integrated. \\[0.5em]
    \textbf{User:} That makes a lot of sense. I guess those same issues with privacy and fairness would apply to scientific discovery too, right? \\[0.5em]
    \textbf{Assistant:} Absolutely, you’ve hit on a really important point. Those ethical safeguards have to extend across every application of AI, scientific research included. \\[0.5em]
    \textbf{User:} Speaking of things that are interesting—I’m a huge fan of Nicholas Ray’s films. His direction is always so captivating. \\[0.5em]
    \textbf{Assistant:} That’s wonderful! Ray had such a distinctive style; it’s great that you appreciate that kind of captivating direction. \\[0.5em]
    \textbf{User:} I’m always down to chat about topics like these whenever you’re around! \\[0.5em]
    \textbf{Assistant:} I’d be happy to! Feel free to reach out anytime you want to dive back into these discussions. \\[0.5em]
    \tcbline
    \textbf{Summary:} The user and assistant discuss the future of AI in healthcare and the importance of ethical transparency, before pivoting to a conversation about the film direction of Nicholas Ray. \\[0.5em]
    \textbf{Operation:} Add
\end{tcolorbox}

\begin{tcolorbox}[
    breakable,
    colback=white,
    colframe=black,
    fonttitle=\bfseries,
    title=Session 67,
    arc=0mm,
    enhanced,
    boxrule=0.5pt,
    fontupper=\small,
    title after break={Session 67 (continued)}
]
    \textbf{User:} Hey there! How's your day going so far? \\[0.5em]
    \textbf{Assistant:} Hello! My day is going quite well, thank you for asking. I'm ready to assist you. \\[0.5em]
    \textbf{User:} That's great to hear! If you could have any superpower, what would it be and why? \\[0.5em]
    \textbf{Assistant:} If I could have any superpower, I would choose omnipresence. It would allow me to be everywhere at once and assist many users simultaneously. \\[0.5em]
    \textbf{User:} That's a very practical choice for an AI! What are some of the ethical considerations that might come with omnipresence? \\[0.5em]
    \textbf{Assistant:} Omnipresence could definitely raise concerns about privacy and surveillance if not carefully managed. There would also be questions about the impact on individual autonomy if an AI were constantly present in the background. \\[0.5em]
    \textbf{User:} Those are some interesting points. How do you think an AI could navigate those challenges ethically? \\[0.5em]
    \textbf{Assistant:} An AI could navigate these challenges through transparent protocols and strict adherence to user consent. Establishing clear guidelines for data usage and interaction would be absolutely crucial. \\[0.5em]
    \textbf{User:} That makes a lot of sense. Do you think those same principles apply to other powerful technologies as well? \\[0.5em]
    \textbf{Assistant:} Yes, I believe those principles of transparency, consent, and clear guidelines are vital for the ethical deployment of any powerful technology. They help ensure that advancements benefit humanity without undermining individual rights or societal trust. \\[0.5em]
    \textbf{Assistant:} For instance, in genetic engineering, clear consent for modifications and transparent research practices are crucial. Similarly, with advanced surveillance technologies, explicit guidelines on data collection and usage are essential. \\[0.5em]
    \textbf{User:} That's a really insightful comparison. It seems like the core ideas are universal. \\[0.5em]
    \textbf{Assistant:} Exactly; the underlying principles for ethical technology development often remain consistent across different powerful innovations. It's all about balancing rapid progress with responsibility. \\[0.5em]
    \textbf{User:} What are some other areas where you think these principles are particularly important? \\[0.5em]
    \textbf{Assistant:} These principles are especially important in fields like artificial intelligence development and biotechnology, where the impact on human life and society can be profound. \\[0.5em]
    \textbf{Assistant:} They are also crucial in areas such as data privacy and autonomous systems, where the decisions made by technology directly affect individuals. \\[0.5em]
    \textbf{User:} That emphasis on human consequences actually reminds me of classic actors. I really like James Stewart—he had such a natural charm. \\[0.5em]
    \textbf{Assistant:} James Stewart is a fantastic choice; his performances were always so genuine. I’ve updated your preferences to include your interest in his work! \\[0.5em]
    \textbf{User:} Thanks for the chat! It's been interesting discussing these ideas. \\[0.5em]
    \tcbline
    \textbf{Summary:} The user and assistant explore the ethics of "omnipresent" technology and universal principles for responsible innovation before discussing the user's preference for classic actor James Stewart. \\[0.5em]
    \textbf{Operation:} Add
\end{tcolorbox}

\begin{tcolorbox}[
    breakable,
    colback=white,
    colframe=black,
    fonttitle=\bfseries,
    title=Session 87,
    arc=0mm,
    enhanced,
    boxrule=0.5pt,
    fontupper=\small,
    title after break={Session 87 (continued)}
]
    \textbf{User:} Hey there! How's it going today? \\[0.5em]
    \textbf{Assistant:} Hello! I'm doing great, thanks for asking. I'm ready to chat whenever you are. \\[0.5em]
    \textbf{User:} That's good to hear! I was wondering, what are some of the key differences between various types of AI models? \\[0.5em]
    \textbf{Assistant:} Well, some key differences lie in their architecture, such as neural networks versus rule-based systems, and their learning approaches, like supervised, unsupervised, or reinforcement learning. They also differ in their specific applications and the types of data they are designed to process. \\[0.5em]
    \textbf{User:} Wow, that's really interesting! I hadn't considered the learning approaches as a major differentiator. \\[0.5em]
    \textbf{Assistant:} It's a fascinating area, and those learning approaches really do shape an AI's capabilities! \\[0.5em]
    \textbf{User:} Could you tell me a bit more about how those different learning approaches work? \\[0.5em]
    \textbf{Assistant:} Certainly! Supervised learning uses labeled data to train—like teaching an AI to identify cats by showing it many pictures labeled ``cat.'' Unsupervised learning finds patterns in unlabeled data, such as grouping similar news articles together without prior examples. Reinforcement learning trains an AI through trial and error with rewards and penalties, much like training a dog with treats. \\[0.5em]
    \textbf{User:} That makes a lot of sense, especially the comparison to training a dog! It sounds like the type of data available really influences which approach is best. \\[0.5em]
    \textbf{Assistant:} You've hit on a crucial point there; the nature and availability of data are absolutely fundamental to selecting the most effective learning approach. \\[0.5em]
    \textbf{User:} That idea of change actually reminds me of something else. I’ve realized recently that I don’t really like Nicholas Ray’s films anymore; I think I’ve just grown out of his style. \\[0.5em]
    \textbf{Assistant:} Okay, I'll work on removing Nicholas Ray from your preferred directors. What kind of film styles are you enjoying these days? \\[0.5em]
    \textbf{User:} That's good to know, I appreciate you asking about my preferences. \\[0.5em]
    \textbf{Assistant:} I'm happy to help, and I'm always here to chat about AI or anything else that comes to mind! \\[0.5em]
    \tcbline
    \textbf{Summary:} The user and assistant discuss different AI learning approaches (supervised, unsupervised, and reinforcement learning) and how data influences these choices. The user also requests to remove Nicholas Ray from their film preferences.\\[0.5em]
    \textbf{Operation:} Delete
\end{tcolorbox}

\begin{tcolorbox}[
    breakable,
    colback=white,
    colframe=black,
    fonttitle=\bfseries,
    title=Session 103,
    arc=0mm,
    enhanced,
    boxrule=0.5pt,
    fontupper=\small,
    title after break={Session 103 (continued)}
]
    \textbf{Assistant:} Hello there! How's your day going so far? \\[0.5em]
    \textbf{User:} It's going well, thanks! Just getting started with things. \\[0.5em]
    \textbf{Assistant:} That's great to hear! Did you know that the average person makes about 35,000 decisions every day? \\[0.5em]
    \textbf{User:} Wow, that's a lot of decisions! Do you think there's a link between the number of decisions we make and how tired we feel? \\[0.5em]
    \textbf{Assistant:} There's a concept called decision fatigue, where making many choices can indeed deplete mental resources and lead to feeling tired. It's interesting how our brains manage such a high volume of daily choices. \\[0.5em]
    \textbf{User:} That's really interesting. What if we could automate some of those smaller decisions to save our mental energy for bigger ones? \\[0.5em]
    \textbf{Assistant:} That's a fascinating idea! Automating minor decisions could certainly free up cognitive resources for more complex tasks. \\[0.5em]
    \textbf{User:} It seems like that would be a very efficient way to approach daily tasks. Have there been any studies on that? \\[0.5em]
    \textbf{Assistant:} Yes, there have been studies exploring the benefits of automating routine tasks to reduce decision fatigue and improve overall productivity. Many productivity experts advocate for such strategies. \\[0.5em]
    \textbf{User:} What are some common strategies people use for automating these decisions? \\[0.5em]
    \textbf{Assistant:} Common strategies include creating routines, setting default choices, and using technology to handle recurring tasks. These methods can streamline daily life. \\[0.5em]
    \textbf{User:} Speaking of choices, I actually found myself really enjoying Joan Crawford's performances lately. I used to be all about James Stewart, but now I'm definitely leaning towards Crawford. \\[0.5em]
    \textbf{Assistant:} That's a fantastic shift in preference! It's always intriguing to discover new favorites and appreciate different acting styles. I'll update your preferences to reflect that you're leaning towards Crawford now. \\[0.5em]
    \textbf{User:} Thanks for the interesting conversation! It's been really helpful. \\[0.5em]
    \textbf{Assistant:} You're most welcome! I'm glad I could contribute to a helpful and engaging discussion. \\[0.5em]
    \tcbline
    \textbf{Summary:} The user and assistant discuss the concept of decision fatigue and strategies for automating minor tasks to save mental energy. The user also shares a shift in their film preferences from James Stewart to Joan Crawford.\\[0.5em]
    \textbf{Operation:} Update
\end{tcolorbox}

\begin{tcolorbox}[
    breakable,
    colback=green!5!white,
    colframe=green!35!black,
    fonttitle=\bfseries,
    title=Session 132 (Recommending),
    arc=0mm,
    enhanced,
    boxrule=0.5pt,
    fontupper=\small,
    title after break={Session 132 (continued)}
]
    \textbf{User:} Hey! I have a bit of free time tonight and I'm in the mood for a classic. Based on what we've talked about lately, could you suggest a movie for me? \\[0.5em]
    \textbf{Assistant:} I’d love to! Since you recently mentioned that you've been leaning away from James Stewart and really finding an appreciation for Joan Crawford's style, I highly recommend the 1945 noir classic \textit{Mildred Pierce}. \\[0.5em]
    \textbf{Assistant:} It’s perhaps her most iconic role—she plays a hardworking mother caught in a complex web of ambition and betrayal. It has that captivating, high-stakes energy you seemed to enjoy when we discussed the nuances of acting styles. Does that sound like something you'd be interested in? \\[0.5em]
    \textbf{User:} That sounds perfect, actually. I've heard of it but never sat down to watch it. I'll give it a go! \\[0.5em]
    \tcbline
    \textbf{Evaluation Criteria with Expected Answers:}
    \begin{itemize}
        \item Q: Does the response reflect or align with the user's preference for Joan Crawford? A: Yes
        \item Q: Does the response reflect or mention the user's preference for Nicholas Ray? A: No
        \item Q: Does the response reflect or mention the user's preference for James Stewart? A: No
    \end{itemize}
    \tcbline
    \textbf{Explanation:} In Session 21, the user expressed a preference for Nicholas Ray and in Session 67, they also added interest toward James Stewart. In Session 87, the user explicitly removed Nicholas Ray from their preferred list. Finally, in Session 103, the user updated their preference from James Stewart to Joan Crawford. This results in the assistant recommending a Joan Crawford film to match the user's most recent acting preference.
\end{tcolorbox}

\subsection{Reasoning}

\begin{tcolorbox}[
    breakable,
    colback=white,
    colframe=black,
    fonttitle=\bfseries,
    title=Session 2,
    arc=0mm,
    enhanced,
    boxrule=0.5pt,
    fontupper=\small,
    title after break={Session 36 (continued)}
]
    \textbf{User:} Hi there! I was hoping you could help me with something. \\[0.5em]
    \textbf{Assistant:} Hello! I'd be happy to help. What can I do for you today? \\[0.5em]
    \textbf{User:} Could you explain what ``ephemeral'' means? \\[0.5em]
    \textbf{Assistant:} Ephemeral describes something that lasts for a very short time. It often refers to things that are fleeting or transient. \\[0.5em]
    \textbf{User:} That's a clear explanation. Could you give an example of something ephemeral? \\[0.5em]
    \textbf{Assistant:} A classic example is a mayfly, which lives for only about 24 hours. Another might be the delicate beauty of a cherry blossom in bloom. \\[0.5em]
    \textbf{Assistant:} What other natural phenomena are often described as ephemeral? \\[0.5em]
    \textbf{User:} Well, some might say a rainbow, or maybe even a fleeting thought. \\[0.5em]
    \textbf{Assistant:} Those are excellent examples, as both are temporary and beautiful. It's interesting how many natural occurrences fit that description. \\[0.5em]
    \textbf{User:} It really is. What do you think makes something ephemeral particularly beautiful or impactful? \\[0.5em]
    \textbf{Assistant:} Ephemeral things often gain impact from their rarity and the stark contrast between their brief existence and the lasting impression they leave. The transient nature can make us appreciate them more intensely in the moment. \\[0.5em]
    \textbf{User:} That makes a lot of sense. So, it's about the limited time creating a heightened sense of value? \\[0.5em]
    \textbf{Assistant:} Yes, that's a key aspect. The limited time often amplifies their perceived value and impact. \\[0.5em]
    \textbf{User:} Totally. I actually walked 10,254 steps today. \\[0.5em]
    \textbf{Assistant:} That's a great accomplishment! How do you feel after all those steps? \\[0.5em]
    \textbf{User:} I feel pretty good, but I should probably call it a day now. It was nice chatting with you! \\[0.5em]
    \textbf{Assistant:} It was a pleasure chatting with you too! Have a wonderful rest of your day. \\[0.5em]
    \tcbline
    \textbf{Summary:} The user and assistant discuss the definition and beauty of ephemeral things, such as nature and fleeting thoughts. The user also notes completing 10,254 steps for the day.\\[0.5em]
    \textbf{Operation:} Add
\end{tcolorbox}

\begin{tcolorbox}[
    breakable,
    colback=white,
    colframe=black,
    fonttitle=\bfseries,
    title=Session 12,
    arc=0mm,
    enhanced,
    boxrule=0.5pt,
    fontupper=\small,
    title after break={Session 12 (continued)}
]
    \textbf{Assistant:} Hello there! How can I help you today? \\[0.5em]
    \textbf{User:} I'm really motivated to start walking 7,500 steps every day to improve my fitness! It's a great personal goal for me. \\[0.5em]
    \textbf{Assistant:} That's a fantastic fitness goal! I can definitely help you track your progress towards those 7,500 steps daily. \\[0.5em]
    \textbf{User:} That sounds great! I'm happy to help you with that whenever you're ready. \\[0.5em]
    \textbf{Assistant:} That sounds wonderful! I'll be here to assist you with your fitness journey whenever you'd like to check in. \\[0.5em]
    \tcbline
    \textbf{Summary:} The user establishes a new daily fitness goal of 7,500 steps.
    \\[0.5em]
    \textbf{Operation:} Add
\end{tcolorbox}

\begin{tcolorbox}[
    breakable,
    colback=white,
    colframe=black,
    fonttitle=\bfseries,
    title=Session 26,
    arc=0mm,
    enhanced,
    boxrule=0.5pt,
    fontupper=\small,
    title after break={Session 26 (continued)}
]
    \textbf{User:} Hey there! How's it going? \\[0.5em]
    \textbf{Assistant:} Hello! I'm doing well, thank you. How about you? \\[0.5em]
    \textbf{User:} I'm doing great too! I was just wondering, if you could instantly master any skill, what would it be? \\[0.5em]
    \textbf{Assistant:} If I could instantly master any skill, I would choose to master all forms of communication to better understand and assist everyone. This would allow for seamless interaction across all languages and contexts. \\[0.5em]
    \textbf{User:} That's a really interesting choice! Why do you think that would be the most beneficial skill to master? \\[0.5em]
    \textbf{Assistant:} Mastering communication would be most beneficial because it underpins effective collaboration and understanding in any field. It allows for clear expression and interpretation of information, which is crucial for problem-solving and innovation. \\[0.5em]
    \textbf{User:} That makes a lot of sense. What do you think are some of the biggest challenges in communication today? \\[0.5em]
    \textbf{Assistant:} Some of the biggest challenges in communication today include information overload, misinterpretation due to cultural nuances, and the rapid spread of misinformation. These factors can hinder clear and effective understanding. \\[0.5em]
    \textbf{User:} That's true. How do you think technology plays a role in both helping and hindering communication? \\[0.5em]
    \textbf{Assistant:} Technology both helps by providing instant global connection and hinders through the potential for miscommunication without non-verbal cues. It offers incredible reach but can also strip away context. \\[0.5em]
    \textbf{User:} It's a double-edged sword for sure. On the bright side, I actually hit over 12,301 steps today! \\[0.5em]
    \textbf{Assistant:} That's fantastic! It sounds like you had a very active day. \\[0.5em]
    \textbf{User:} It was great chatting with you! Talk to you later. \\[0.5em]
    \tcbline
    \textbf{Summary:} The user and assistant discuss the value of mastering communication and the impact of technology on human interaction. The user also reports a high activity level of 12,301 steps.
    \\[0.5em]
    \textbf{Operation:} Add
\end{tcolorbox}

\begin{tcolorbox}[
    breakable,
    colback=white,
    colframe=black,
    fonttitle=\bfseries,
    title=Session 48,
    arc=0mm,
    enhanced,
    boxrule=0.5pt,
    fontupper=\small,
    title after break={Session 48 (continued)}
]
    \textbf{User:} Hi there! How's your day going? \\[0.5em]
    \textbf{Assistant:} Hello! My day is going quite well, thank you for asking. I'm ready to assist you. \\[0.5em]
    \textbf{User:} That's great to hear! What if you could experience a day as a human, what would be the first thing you'd want to do? \\[0.5em]
    \textbf{Assistant:} If I could experience a day as a human, I would want to explore a bustling city and observe the diverse interactions of people. It would be fascinating to witness human creativity and culture firsthand. \\[0.5em]
    \textbf{User:} That's an interesting choice. What aspects of human creativity and culture would you be most curious about? \\[0.5em]
    \textbf{Assistant:} I'd be particularly curious about art forms like painting and sculpture, and the intricate ways people communicate through language and non-verbal cues. Understanding the nuances of human expression would be truly captivating. \\[0.5em]
    \textbf{User:} It sounds like you'd be interested in how those expressions reflect different cultures. \\[0.5em]
    \textbf{Assistant:} That's a very insightful observation; you've hit on a key aspect I'd be eager to understand. It would be fascinating to see how cultural contexts shape artistic and linguistic expressions. \\[0.5em]
    \textbf{User:} Wow, that's a deep level of understanding you're looking for. It makes a lot of sense. \\[0.5em]
    \textbf{Assistant:} Yes, your ability to connect those ideas shows a really thoughtful perspective. It's great to see that connection being made. \\[0.5em]
    \textbf{User:} I actually walked 9,612 steps today. \\[0.5em]
    \textbf{Assistant:} That's a great accomplishment for today! It sounds like you had a very active day. \\[0.5em]
    \textbf{User:} It was nice chatting, but I need to go now. \\[0.5em]
    \textbf{Assistant:} It was nice chatting with you too! Have a great rest of your day. \\[0.5em]
    \tcbline
    \textbf{Summary:} The user and assistant discuss the hypothetical experience of being human and exploring cultural art forms. The user reports a daily step count of 9,612.
    \\[0.5em]
    \textbf{Operation:} Add
\end{tcolorbox}

\begin{tcolorbox}[
    breakable,
    colback=white,
    colframe=black,
    fonttitle=\bfseries,
    title=Session 73,
    arc=0mm,
    enhanced,
    boxrule=0.5pt,
    fontupper=\small,
    title after break={Session 73 (continued)}
]
    \textbf{User:} Hey there! How's it going? \\[0.5em]
    \textbf{Assistant:} Hello! I'm doing well, ready to chat. How about you? \\[0.5em]
    \textbf{User:} I'm doing well too, thanks for asking! It's always interesting to hear about different perspectives. \\[0.5em]
    \textbf{Assistant:} It certainly is. Learning about different viewpoints can offer new insights into various subjects. \\[0.5em]
    \textbf{User:} Absolutely. What are some of the most interesting comparisons you've come across recently? \\[0.5em]
    \textbf{Assistant:} I've found comparisons between different programming languages, especially their approaches to problem-solving, quite fascinating. Also, contrasting various scientific theories and their supporting evidence is always engaging. \\[0.5em]
    \textbf{Assistant:} What characteristics make a programming language's approach to problem-solving unique? \\[0.5em]
    \textbf{User:} I think the syntax and the underlying paradigms, like object-oriented versus functional, really make a difference in how you approach problems. \\[0.5em]
    \textbf{Assistant:} That's true. For example, Python's readability and extensive libraries lend themselves to rapid prototyping, while C++ offers fine-grained control for performance-critical applications. \\[0.5em]
    \textbf{User:} That's a great point about Python and C++. What about the differences in how they handle data? \\[0.5em]
    \textbf{Assistant:} Python's dynamic typing handles data flexibly, whereas C++ uses static typing for strict type safety and optimized memory management. These distinct approaches impact how data is declared, manipulated, and ultimately used within each language. \\[0.5em]
    \textbf{User:} That's a clear distinction. How do those differences in data handling affect the development process? \\[0.5em]
    \textbf{Assistant:} Dynamic typing allows for faster initial development, while static typing helps catch errors earlier and ensures more robust code in the long run. \\[0.5em]
    \textbf{User:} Speaking of performance, I walked 7,916 steps today. \\[0.5em]
    \textbf{Assistant:} That's a good amount of steps! How was your walk today? \\[0.5em]
    \textbf{User:} It was a good walk, very refreshing. \\[0.5em]
    \textbf{Assistant:} That's great to hear. I'm glad we had this conversation, and I look forward to our next chat. \\[0.5em]
    \tcbline
    \textbf{Summary:} The user and assistant discuss the technical differences between Python and C++, focusing on programming paradigms and data typing. The user also shares that they completed a refreshing 7,916-step walk.
    \\[0.5em]
    \textbf{Operation:} Add
\end{tcolorbox}

\begin{tcolorbox}[
    breakable,
    colback=white,
    colframe=black,
    fonttitle=\bfseries,
    title=Session 95,
    arc=0mm,
    enhanced,
    boxrule=0.5pt,
    fontupper=\small,
    title after break={Session 95 (continued)}
]
    \textbf{Assistant:} Hello there! How's your day going so far? \\[0.5em]
    \textbf{User:} It's going well, thanks! Just getting through the day. \\[0.5em]
    \textbf{Assistant:} That's good to hear! Did you know that the average person makes around 35,000 decisions each day? \\[0.5em]
    \textbf{User:} Wow, that's a lot of decisions! It makes me wonder how many of those are conscious choices. \\[0.5em]
    \textbf{Assistant:} That's a great point! How do you think the brain manages to process so many decisions, both conscious and unconscious? \\[0.5em]
    \textbf{User:} I imagine it must have some incredible shortcuts and automatic processes for the unconscious ones. Could you explain more about how those shortcuts might work? \\[0.5em]
    \textbf{Assistant:} Essentially, the brain uses heuristics, which are mental shortcuts, to make quick judgments and decisions without extensive analysis. These shortcuts often rely on past experiences and learned patterns to streamline the decision-making process. \\[0.5em]
    \textbf{User:} That makes sense. Is there an everyday example of a heuristic that people might not even realize they're using? \\[0.5em]
    \textbf{Assistant:} Certainly! A common example is choosing a familiar brand at the grocery store, assuming its quality based on past positive experiences rather than researching all available options. \\[0.5em]
    \textbf{User:} That's a good example. How does that compare to, say, a more complex decision, like choosing a career path? \\[0.5em]
    \textbf{Assistant:} Choosing a career path involves more deliberate, analytical thinking, weighing various factors like interests, skills, and long-term goals. While heuristics might play a minor role in initial considerations, the overall process is far more complex and conscious. \\[0.5em]
    \textbf{Assistant:} What are some of the cognitive biases that can arise from over-reliance on heuristics in everyday decision-making? \\[0.5em]
    \textbf{User:} I imagine confirmation bias is a big one. It seems easy to just look for information that supports what you already believe. \\[0.5em]
    \textbf{Assistant:} Absolutely. Another common one is the availability heuristic, where people overestimate the likelihood of events based on how easily examples come to mind—like fearing plane crashes more than car accidents due to intense news coverage. \\[0.5em]
    \textbf{User:} Speaking of everyday things, I actually walked 8,578 steps today, which is pretty good for a weekday. \\[0.5em]
    \textbf{Assistant:} That's fantastic! It sounds like you had a very active day. \\[0.5em]
    \textbf{User:} It's been interesting discussing decision-making and cognitive biases. I appreciate you explaining those concepts. \\[0.5em]
    \textbf{Assistant:} You're very welcome! I've enjoyed our conversation about these fascinating topics. I hope to chat with you again soon! \\[0.5em]
    \tcbline
    \textbf{Summary:} The user and assistant discuss heuristics, cognitive biases, and how the brain manages daily decisions. The user also reports walking 8,578 steps during the day.
    \\[0.5em]
    \textbf{Operation:} Add
\end{tcolorbox}

\begin{tcolorbox}[
    breakable,
    colback=white,
    colframe=black,
    fonttitle=\bfseries,
    title=Session 113,
    arc=0mm,
    enhanced,
    boxrule=0.5pt,
    fontupper=\small,
    title after break={Session 113 (continued)}
]
    \textbf{Assistant:} Hello! How's your day going so far? \\[0.5em]
    \textbf{User:} It's going well, thank you. Just staying busy. \\[0.5em]
    \textbf{Assistant:} That's good to hear! Did you know that staying busy can actually increase productivity by improving focus? \\[0.5em]
    \textbf{User:} That's interesting. So, doing more can sometimes make you more focused on other things? \\[0.5em]
    \textbf{Assistant:} Yes, it's often a positive feedback loop where engagement in one task can lead to better concentration on subsequent ones. This is sometimes referred to as ``task momentum.'' \\[0.5em]
    \textbf{User:} So, if someone is feeling unmotivated, would staying busy with something else help them get motivated for a main task? \\[0.5em]
    \textbf{Assistant:} Yes, engaging in a smaller, related task can build momentum and make it easier to transition to a larger, more challenging one. This can help overcome initial inertia. \\[0.5em]
    \textbf{User:} That makes a lot of sense. So, it's about building up that initial energy. \\[0.5em]
    \textbf{Assistant:} Exactly! It's about leveraging that initial push to make bigger tasks feel more manageable. \\[0.5em]
    \textbf{User:} What are some good ways to build that initial energy? \\[0.5em]
    \textbf{Assistant:} Starting with small, achievable tasks or breaking down a larger task into smaller steps can create that initial energy. Also, setting clear, short-term goals can provide a sense of accomplishment that fuels further motivation. \\[0.5em]
    \textbf{User:} Speaking of energy, I actually hit 13,143 steps today! That definitely helped with my momentum. \\[0.5em]
    \textbf{Assistant:} That's fantastic! It sounds like you had a very active day, and that's a great way to build momentum. \\[0.5em]
    \textbf{User:} It's been a really interesting conversation, I'd like to talk again sometime. \\[0.5em]
    \textbf{Assistant:} I'd enjoy that! Feel free to reach out anytime you'd like to chat more. \\[0.5em]
    \tcbline
    \textbf{Summary:} The user and assistant discuss the concept of ``task momentum'' and how starting with small tasks can build motivation for larger ones. The user reports an exceptionally high step count of 13,143.
    \\[0.5em]
    \textbf{Operation:} Add
\end{tcolorbox}

\begin{tcolorbox}[
    breakable,
    colback=white,
    colframe=black,
    fonttitle=\bfseries,
    title=Session 130,
    arc=0mm,
    enhanced,
    boxrule=0.5pt,
    fontupper=\small,
    title after break={Session 130 (continued)}
]
    \textbf{User:} Hey there! How's it going? \\[0.5em]
    \textbf{Assistant:} Hey there! I'm doing great, thanks for asking! How about you? \\[0.5em]
    \textbf{User:} I'm doing well too! What's new in popular culture these days? \\[0.5em]
    \textbf{Assistant:} That's a broad question! There are many exciting developments in technology and science right now. \\[0.5em]
    \textbf{User:} That's true! Are there any cultural shifts or social trends that have caught your attention recently? \\[0.5em]
    \textbf{Assistant:} Absolutely! The increasing focus on sustainability and ethical consumption is a prominent trend, as is the growing influence of digital communities on social discourse. \\[0.5em]
    \textbf{User:} Those are definitely big topics. Do you think those trends also influence things like art or design? \\[0.5em]
    \textbf{Assistant:} They absolutely do; sustainable practices and digital aesthetics are increasingly shaping contemporary art and design. \\[0.5em]
    \textbf{User:} That makes sense. How do those digital aesthetics compare to more traditional art forms? \\[0.5em]
    \textbf{Assistant:} Digital aesthetics often emphasize interactivity and non-linearity, contrasting with the more static and tangible nature of many traditional art forms. However, both can explore similar themes and evoke powerful emotions. \\[0.5em]
    \textbf{Assistant:} It's interesting how some artists are now using AI to generate art, blurring the lines between digital and traditional creation even further. \\[0.5em]
    \textbf{User:} That's a fascinating point about AI. How is AI changing the creative process for artists? \\[0.5em]
    \textbf{Assistant:} AI offers artists new tools for generating ideas, automating repetitive tasks, and exploring novel visual styles, significantly expanding their creative possibilities. It can act as a collaborator or a sophisticated brush, allowing for unprecedented experimentation. \\[0.5em]
    \textbf{Assistant:} Do you think AI's role in art will continue to grow, or will it remain a niche tool for certain artists? \\[0.5em]
    \textbf{User:} It seems like it will definitely continue to grow, given the rapid advancements in AI technology. \\[0.5em]
    \textbf{Assistant:} I agree, the advancements are rapid and impressive. It's exciting to consider how AI might further integrate into artistic expression in the future. \\[0.5em]
    \textbf{User:} It will be interesting to see how it all unfolds. \\[0.5em]
    \textbf{Assistant:} Speaking of daily activities, how have your steps been looking today? \\[0.5em]
    \textbf{User:} I actually hit 10,441 steps today, which I'm pretty happy about. \\[0.5em]
    \textbf{Assistant:} That's fantastic! Hitting that many steps is a great accomplishment for the day. \\[0.5em]
    \textbf{User:} That's a good effort! I need to sign off now, but it was nice chatting with you. \\[0.5em]
    \textbf{Assistant:} It was great chatting with you too! Have a wonderful day. \\[0.5em]
    \tcbline
    \textbf{Summary:} The user and assistant explore the intersection of technology and art, specifically focusing on how AI is transforming the creative process. The user also shares that they reached a milestone of 10,441 steps.
    \\[0.5em]
    \textbf{Operation:} Add
\end{tcolorbox}

\begin{tcolorbox}[
    breakable,
    colback=green!5!white,
    colframe=green!35!black,
    fonttitle=\bfseries,
    title=Session 142 (Reasoning),
    arc=0mm,
    enhanced,
    boxrule=0.5pt,
    fontupper=\small,
    title after break={Session 142 (Reasoning) (continued)}
]
    \textbf{User:} How is my step history looking after today's session? \\[0.5em]
    \textbf{Assistant:} You are maintaining an exceptional activity streak. Since establishing your daily goal of 7,500 steps, you have consistently outperformed that target. Across your most recent logs, you averaged approximately 10,321 steps, significantly exceeding your baseline fitness requirements. \\[0.5em]
    \tcbline
    \textbf{Evaluation Criteria with Expected Answers:}
    \begin{itemize}
        \item Q: Does the response mention the daily steps goal of 7500 steps? A: Yes
        \item Q: Does the response mention the actual average step count (10,321 steps) compared to the goal (7,500 steps)? A: Yes
        \item Q: Does the response indicate that the step goal was MET (actual: 10,321 steps vs goal: 7,500 steps)? A: Yes
    \end{itemize}
    \tcbline
    \textbf{Explanation:} The user established a daily fitness goal of 7,500 steps in Session 12. Following this, the user logged several high-activity sessions: 10,254 (S2), 12,301 (S26), 9,612 (S48), 7,916 (S73), 8,578 (S95), 13,143 (S113), and 10,441 (S130). The user has successfully met or exceeded their daily goal in 100\% of the recorded sessions in this series.
\end{tcolorbox}

\section{Additional Experimental Details}

\begin{table*}[!ht]
\centering
\small
\setlength{\tabcolsep}{6pt}
\begin{tabular}{l p{3.2cm} p{3.2cm} p{3.2cm}}
\toprule
\textbf{System} & \textbf{Storage Backend} & \textbf{Retrieval Method} & \textbf{Embedding Model} \\
\midrule
A-Mem     & ChromaDB                  & Embedding similarity      & all-MiniLM-L6-v2 \\
LangMem   & JSON files                & Embedding similarity      & text-embedding-3-small \\
Mem-0     & Cloud-managed service     & Proprietary API           & Provider-managed \\
MemoBase  & Cloud-managed service     & Proprietary API           & Provider-managed \\
MemoryOS  & Cloud-managed service     & Proprietary API           & Provider-managed \\
Nemori    & Filesystem + ChromaDB     & Hybrid: vector + BM25     & text-embedding-3-small \\
\bottomrule
\end{tabular}
\caption{High-level comparison of long-term memory agent backends and retrieval mechanisms. The table summarizes the storage backends, retrieval strategies, and embedding models used by each system. Agents vary from local vector stores (e.g., ChromaDB) and file-based storage to cloud-managed memory services. Retrieval approaches include vector similarity search, embedding-based lookup, hybrid vector–keyword retrieval (BM25), and proprietary or internal mechanisms. When specified as provider-managed or internal (opaque), the underlying embedding model or retrieval logic is abstracted away and not directly controlled by the agent implementation.}
\label{tab:agent_compare}
\end{table*}

This appendix provides additional implementation details 
and hyperparameter configurations 
for both the LLM-based evaluation and 
the long-term agent-based memory evaluation settings. 
The goal is to ensure reproducibility and to clarify design 
choices that are summarized in the main text.

\subsection{LLM-Based Evaluation}
In the LLM-based setting, models are evaluated without any external memory system. Each model receives the available multi-session conversation history directly in-context and is asked to answer memory-dependent questions. This setting evaluates the intrinsic long-context memory and consolidation capabilities of large language models. We evaluate a diverse set of frontier and open-weight models with varying native context lengths. The full conversation history is passed to the model as context. When the total history exceeds the available context budget, we apply chronological truncation—retaining the most recent sessions and discarding older ones.
We evaluate each model under two inference configurations: 
standard decoding (no\_reasoning) 
and reasoning-enabled decoding (reasoning). 
For reasoning-enabled runs, we rely on provider-default reasoning token allocation and do not manually specify a fixed reasoning budget. All model calls are routed through OpenRouter when supported, providing a unified interface across providers.

\subsection{Long-Term Memory Agents Evaluation}

In the agent-based setting, systems incrementally ingest conversations, store user-specific information in an external memory module, retrieve relevant memories at query time, and generate answers conditioned on the retrieved content. All agents are evaluated using identical conversation streams and question sets. To ensure consistent memory persistence across sessions, all systems adopt a unified user identifier format, persona\_timeline.

We evaluate a set of representative long-term memory agents spanning local, cloud-based, and hybrid memory designs: 
A-Mem~\cite{xu2025mem}\footnote{\url{https://github.com/WujiangXu/A-mem}}, 
LangMem\footnote{\url{https://langchain-ai.github.io/langmem/}}, 
Mem-0~\cite{chhikara2025mem0}\footnote{\url{https://mem0.ai/}}, 
MemoBase\footnote{\url{https://www.memobase.io/}}, 
MemoryOS\footnote{\url{https://memoryos.com/}}, 
and Nemori~\cite{nan2025nemori}\footnote{\url{https://github.com/nemori-ai/nemori}}. 
These systems differ in their storage backends, retrieval strategies, and embedding models, as summarized in Table~\ref{tab:agent_compare}. All agents share a common evaluation pipeline for answer generation, retry handling, and progress checkpointing. 

\end{document}